\pdfoutput=1

\documentclass[11pt]{article}

\usepackage[table,xcdraw]{xcolor}
\usepackage[preprint]{acl}

\usepackage{times}
\usepackage{latexsym}

\usepackage[T1]{fontenc}

\usepackage[utf8]{inputenc}

\usepackage{microtype}

\usepackage{inconsolata}

\usepackage{graphicx}
\usepackage{amsmath}
\usepackage{amssymb}
\usepackage{booktabs}
\usepackage{multirow}
\usepackage{hyperref}

\title{\textsc{K-comp}: Retrieval-Augmented Medical Domain Question Answering With Knowledge-Injected Compressor}

\author{Jeonghun Cho \\
POSTECH GSAI\\
\texttt{jeonghuncho@postech.ac.kr} \\\And
Gary Geunbae Lee\\
POSTECH GSAI\\
POSTECH CSE\\
\texttt{gblee@postech.ac.kr} \\}

\begin{document}
\maketitle

\begin{abstract}
Retrieval-augmented question answering (QA) integrates external information and thereby increases the QA accuracy of reader models that lack domain knowledge. However, documents retrieved for closed domains require high expertise, so the reader model may have difficulty fully comprehending the text. Moreover, the retrieved documents contain thousands of tokens, some unrelated to the question. As a result, the documents include some inaccurate information, which could lead the reader model to mistrust the passages and could result in hallucinations. To solve these problems, we propose \textbf{\textsc{K-comp}} (\textbf{\textsc{K}}nowledge-injected \textbf{\textsc{comp}}ressor) which provides the knowledge required to answer correctly. The compressor automatically generates the prior knowledge necessary to facilitate the answer process prior to compression of the retrieved passages. Subsequently, the passages are compressed autoregressively, with the generated knowledge being integrated into the compression process. This process ensures alignment between the question intent and the compressed context. By augmenting this prior knowledge and concise context, the reader models are guided toward relevant answers and trust the context.\footnote{Our implementation can be accessed at \url{https://github.com/jeonghun3572/K-COMP}.}
\end{abstract}
\section{Introduction}
Retrieval-augmented question answering (QA) is a task where passages related to a question are appended to the prompt such that a reader model can reference them and infer the correct answer~\citep{ahmad-etal-2019-reqa, guo-etal-2021-multireqa}.
\begin{figure}[t!]
  \centering
  \includegraphics[width=\linewidth]{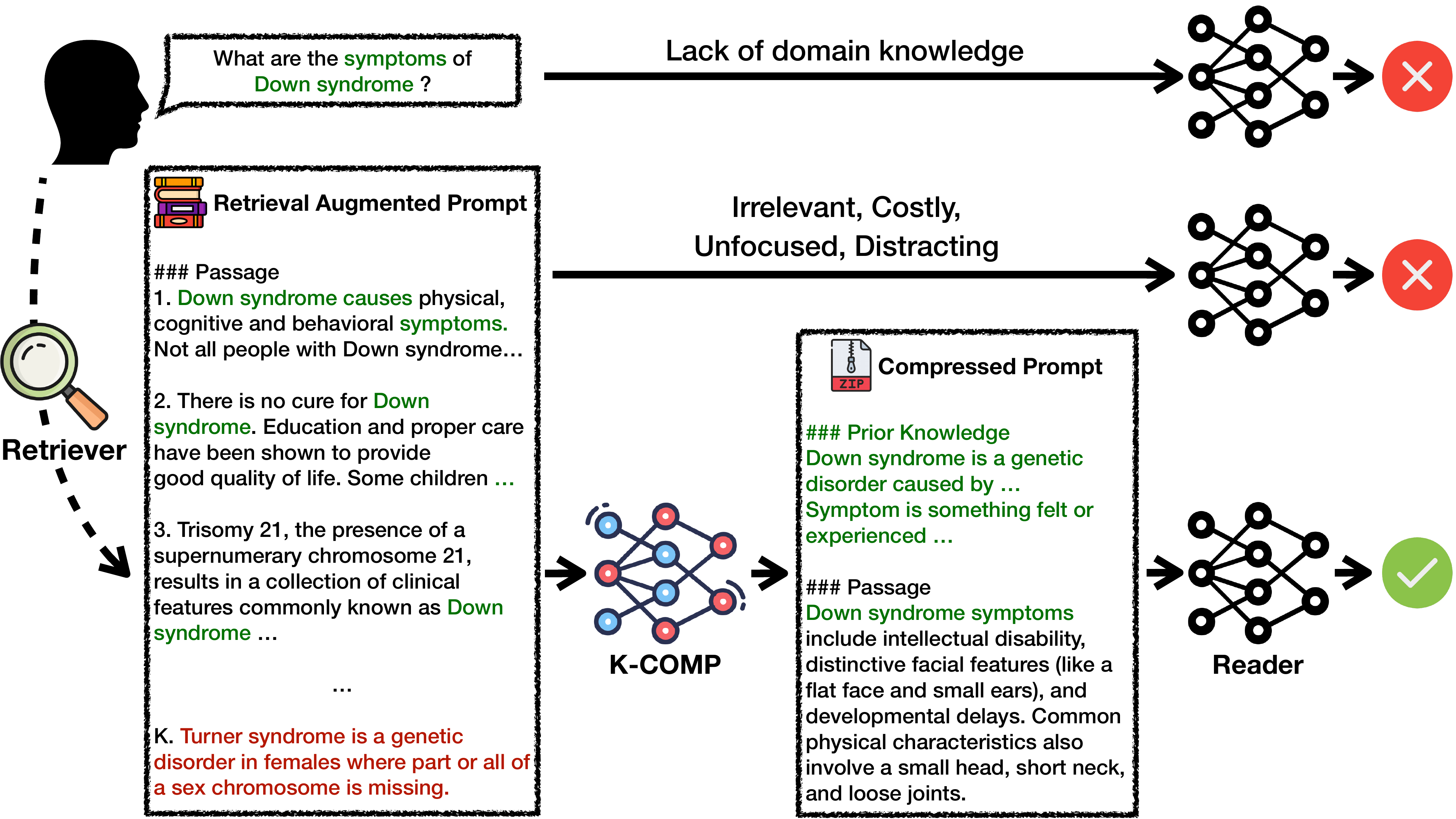} 
  \caption{\textsc{K-comp} helps the reader model infer accurate responses by using domain knowledge and compressed context aligned with the question.}
  \label{fig:main_figure}
\end{figure}
However, several limitations impede retrieval-augmented approaches in closed domains with large language models (LLMs) as readers. First, the documents retrieved for closed domains require domain expertise, so the reader may not trust the whole text. When faced with unfamiliar input, the model exhibits an availability bias toward commonly known knowledge, making it more willing to believe in information it can easily recall~\citep{jin-etal-2024-tug}. Also, retrieved passages contain thousands of tokens and are sometimes unrelated to the question. This can cause the language model to distrust the passages, perceive them as irrelevant noise, and generate answers that do not consider them. These problems lead to hallucinations~\citep{10.1145/3571730}, which result in the model generating inaccurate answers or inferring plausible but false responses. Lastly, LLMs are sensitive to the order of retrieved documents and the prompting method. Specifically, LLMs can have difficulty finding the necessary information within lengthy input prompts, especially when key information or correct answer clues are located in the middle of the prompt~\citep{liu-etal-2024-lost, xu2024retrievalmeetslongcontext}.

To address these issues, we propose \textsc{K-comp} (knowledge-injected compressor). We aim to use an autoregressive LLM as a compressor with the domain knowledge needed to answer the question and increase the alignment of the retrieved passages with the question intent. Additionally, when the compressor is trained in domain-related terms and information, it becomes able to recognize the entities that occur in the question and provide descriptions for them. This process is significant for closed domains that require substantial prior knowledge. For retrieval augmentation, we use a large amount of text from domain-specific sources, including Wikipedia. We exploit the advantages of domain relevance by efficiently reusing it when annotating prior knowledge, not just for retrieval. Furthermore, we use a causal masking objective~\citep{aghajanyan2022cm3causalmaskedmultimodal} during the training phase to inject domain knowledge into the compressor.

In summary, our contributions are as follows:
\begin{itemize}
\item We propose a novel approach to generate knowledge-injected summaries adapted for the medical domain. We incorporate causal masking to inject knowledge into the compressor without modifying its structure. This approach ensures that the summary is aligned with the question.
\item Even without domain knowledge in the reader model, \textsc{K-comp} provides the description of the medical jargon to answer the question, thereby enabling LLMs with diverse backgrounds to handle medical questions more accurately.
\item Experiments on three medical datasets show that \textsc{K-comp} improves performance over other query-based prompt compression methodologies and standard retrieval-augmented generation (RAG) without compression.
\item \textsc{K-comp} has been shown to be effective when applied to previously unseen data, thereby presenting evidence that our method provides additional novel contributions in data-scarce closed domain environments.
\end{itemize}
\section{Related Work}
\paragraph{Text Infilling} Models such as BERT~\citep{devlin-etal-2019-bert}, SpanBERT~\citep{joshi-etal-2020-spanbert}, T5~\citep{JMLR:v21:20-074}, and BART~\citep{lewis-etal-2020-bart}, are pre-trained using masked language modeling within a bidirectional encoder architecture. They have shown strong performance in infilling short and contiguous masked token spans. However, the bidirectional attention mechanism typically restricts the fillable span length to dimensions significantly shorter than a sentence.

In contrast, decoder-only models such as GLM~\citep{du-etal-2022-glm}, CM3~\citep{aghajanyan2022cm3causalmaskedmultimodal}, and InCoder \citep{fried2023incoder} operate by left-to-right generation. They can accommodate variable infill span lengths. Causal masking~\citep{aghajanyan2022cm3causalmaskedmultimodal} or fill-in-the-middle~\citep{bavarian2022efficienttraininglanguagemodels} methods predict masked spans from the posterior context. These methods have their generative capabilities, which increase the length of infill spans. They can also exploit the advantages of considering contextual relationships that surround the masked span. The proposed method has the capability to fill the span by considering bidirectional context, as well as align the generated summary with the question by regressively encoding the infilled span.

\paragraph{Prompt Compression} Several studies have demonstrated that prompt augmentations effectively enhance the performance of LLMs across various tasks~\citep{liu-etal-2023-recap, ram-etal-2023-context, ryu-etal-2023-retrieval, wang2024ratretrievalaugmentedthoughts, long-etal-2023-adapt, yagnik2024medlmexploringlanguagemodels}. Yet, the relevance and reliability of the augmented passages are significant challenges in prompt augmentations. In order to address this issue, recent studies have attempted to extract content from ambiguous and lengthy passages directly. \citet{kim2024sure} eliminates irrelevant information while maximizing the extraction of accurate information, whereas \citet{yang-etal-2023-prca} leverages the black-box LLMs by applying a reward-based method during compressor training to generate summaries. RECOMP~\citep{xu2024recomp} selects and augments the summary with the highest end-task performance by using prompts in which non-essential summaries are set to empty strings if necessary. LLMLingua~\citep{jiang-etal-2023-llmlingua} dynamically assigns different compression rates to various components within the prompt. In contrast, \textsc{K-comp} focuses on the keywords needed to answer the question, emphasizing the alignment between the compressed context and the question.
\section{Causal Knowledge Injection}
Causal models trained using autoregressive language modeling rely exclusively on the context to the left of generated tokens to predict subsequent tokens~\citep{NEURIPS2020_1457c0d6}.
This attribute confers an \textit{\textbf{advantage in causally generating entire documents}}, such as text generation. However, these models show limited proficiency in tasks that require an understanding of post-positional relationships for span infilling. Conversely, masked language models excel at \textit{\textbf{predicting masked spans}} by referencing attention scores from tokens located \textit{\textbf{both anteriorly and posteriorly}}. Nonetheless, their training objective is limited to decoding only short segments of passages~\citep{devlin-etal-2019-bert, joshi-etal-2020-spanbert}.

We are inspired by causal masking~\citep{aghajanyan2022cm3causalmaskedmultimodal} that combines the advantages of both objectives. We focus on the \textit{\textbf{masked medical entities}} within the \textit{\textbf{question (prior context)}} and aim to predict them by considering the \textit{\textbf{retrieved snippets (subsequent context)}}. Subsequently, by \textit{\textbf{auto-regressively compressing the retrieved snippets}}, we can effectively leverage both advantages.
\section{Methods}
In this section, we report our proposed approach for knowledge-injected compression and retrieval augmentation. To retrieve passages similar to a question, we construct a retrieval pipeline composed of a large corpus (\S\ref{sec:Retrieval Framework}). Next, we explain the data processing steps for training (\S\ref{sec:Data processing}). Finally, we detail the training scheme for \textsc{K-comp} with the proposed objective and explain the inference phase for retrieval augmentation (\S\ref{sec:Prompt Compressor}). Figure~\ref{fig:main_figure} shows an overview of the prompts that \textsc{K-comp} consists of.
\subsection{Retrieval Framework}
\label{sec:Retrieval Framework}
Closed domain tasks have not been as thoroughly explored as open domain tasks, which have achieved notable performance enhancements using Wikipedia as a retrieval corpus~\citep{karpukhin-etal-2020-dense}. In contrast to open domains, the challenge in closed domains is that unified corpora have not been established. Research endeavors, such as~\citet{xiong2024benchmarking, wang2024augmentingblackboxllmsmedical}, are currently underway to address this gap. To ensure coverage of both general and domain knowledge, we adopt the MedCorp corpus~\citep{xiong2024benchmarking} as our retrieval corpus. It combines Wikipedia, PubMed\footnote{\label{foot:pubmed}\url{https://pubmed.ncbi.nlm.nih.gov/}}, StatPearls\footnote{\url{https://www.statpearls.com/}}, and textbooks~\citep{app11146421}. As our retriever, we employ embedding-based $k$-NN search~\citep{johnson2019billion} to mitigate bottlenecks and efficiently execute similarity searches on our large-scale corpus comprising four distinct text corpora\footnote{We use Nomic Embed~\citep{nussbaum2024nomicembedtrainingreproducible}.}.

\subsection{Ground-Truth Data}
\label{sec:Data processing}
\paragraph{Entity-Description} We rely on off-the-shelf tools to perform named-entity recognition\footnote{\label{foot:scispacy}We use ScispaCy~\citep{neumann-etal-2019-scispacy} package.}, which identifies biomedical entities $\mathcal{E}=\{e_i\}$ in each question for masking. Retrieval corpus $\mathbb{C}$  is constituted of title and text pairs, with the first sentence of each text assumed to be a short description of the title~\citep{https://doi.org/10.48550/arxiv.2302.09170}. Subsequently, the pairs of titles and short descriptions are matched with the entities and their corresponding knowledge $d_i$. We assume that the questions in the training dataset contain at least one entity. In the absence of an entity in a given question, the data are excluded. Similarly, instances lacking a corresponding title in the retrieval corpus are also filtered out of the training dataset (Table~\ref{tab:dataset_statistics}).

However, in the test data, \textsc{K-comp} unveils a novel contribution by automatically generating domain-specific entity descriptions during inference even when no annotate exists for the entity in question. This obviates the need for costly and unnecessary tasks, such as searching for medical terms or finding definitions within the corpus.

\paragraph{Summary} To synthesize gold summaries $\mathcal{S}$, GPT-4o-mini\footnote{\label{foot:gpt-mini}We use gpt-4o-mini-2024-07-18~\citep{gpt4o-mini}.} compresses the passages by considering $\{\mathcal{P}, \mathcal{E}\}$ input pairs, and the number of passages used for synthesis is set to five, i.e., $|\mathcal{P}|=5$. Notably, we explicitly prohibit the inclusion of the question in the summary synthesis process. This is because incorporating the question into the input prompt for generating the summary may result in a focus shift from the generation of keyword-focused summaries to the formulation of a summary that is aimed at answering the question. Detailed instructions for the summary synthesis are provided in Table~\ref{tab:summary_generation}.

\subsection{\textsc{K-comp}}
\label{sec:Prompt Compressor}
\paragraph{Preliminary} $q=\left[q^1, q^2, ..., q^N \right]$, where $q^N$ represents the $N$-th token in $q$. We use the special token $\texttt{<ent>}$ to mask each medical entity spans within $q$, $q_m = \left[q^1,...,\texttt{<ent>},..., q^{N-l}\right]$. Also, a special $\texttt{<eod>}$ token is appended at the end of the description of the corresponding entity, $d_i=\left[d_i^1,...,d_i^M, \texttt{<eod>}\right]$. An example is provided as follows:
\begin{align*}
    &\begin{aligned}
        q_m = \text{What are the \texttt{<ent>} of \texttt{<ent>}?}
    \end{aligned}\\
    &\begin{aligned}
        d_{1} = \text{symptom: \texttt{\{\{description\}\}<eod>}}
    \end{aligned}\\
    &\begin{aligned}
        d_{2} = \text{Down syndrome: \texttt{\{\{description\}\}<eod>}}
    \end{aligned}
\end{align*}

By concatenating $q_m$ and $\mathcal{P}$ in the correct sequence, the masked spans can be predicted based on the preceding and subsequent context. We define the dataset for the compressor as $(q_m \oplus \mathcal{P}, \mathcal{S},\mathcal{E},\mathcal{D})$, where $\mathcal{D}=\{d_i\}$ and $\mathcal{S}$ is a gold summary.

\paragraph{Training} Given an input query $q_m$ and the set of retrieved passages $\mathcal{P}=\{p_1,p_2, ..., p_5\}$, \textsc{K-comp} aims to train a causal model $f(q_m \oplus \mathcal{P})$ to generate $\mathcal{E}$, $\mathcal{D}$, and then $\mathcal{S}$ auto-regressively.

The compressor is trained to encode $q_m \oplus \mathcal{P}$ to generate \texttt{<ent>} tokens and their corresponding descriptions:
\begin{multline}
\nonumber
    P_{\theta}(\mathcal{E},\mathcal{D} | q_m \oplus \mathcal{P}) \\
    = \prod_{i} \left(\prod_{\alpha,\beta} P_\theta(e_i^{\alpha}, d_i^{\beta} | e_i^{<\alpha},d_i^{<\beta}, q_m \oplus \mathcal{P}) \right)
\end{multline}
where $\theta$ represents the parameters of \textsc{K-comp}.

This approach facilitates the incorporation of descriptions into the prompt for the reader model and ensures that the generated entities and their descriptions are regressively encoded. As a result, a summary is generated in a causal manner with attention to the entities within the question and their related knowledge, thereby composing a summary centered on these domain entities.
\begin{multline}
\nonumber
    P_{\theta}(s | \mathcal{E},\mathcal{D},q_m \oplus \mathcal{P}) \\
    = \prod_{\gamma} P_\theta(s^{\gamma} | s^{<\gamma},\mathcal{E},\mathcal{D},q_m \oplus \mathcal{P})
\end{multline}
We fine-tuned the compressor using the standard next token prediction with cross-entropy loss:
\begin{multline}
\nonumber
    P_\theta(\mathcal{E},\mathcal{D},s | q_m \oplus \mathcal{P}) \\
    = P_\theta(\mathcal{E},\mathcal{D} | q_m \oplus \mathcal{P}) \times P_\theta(s | \mathcal{E},\mathcal{D},q_m \oplus \mathcal{P})
\end{multline}
\begin{equation}
\nonumber
    \therefore L(\theta) = -\mathbb{E}(\log P_\theta(\mathcal{E}, \mathcal{D}, s \mid q_m \oplus \mathcal{P}))
\end{equation}

\paragraph{Inference} At inference time, documents are retrieved in advance to construct the compressor input batch $\{\mathcal{P}, q_m\}$. This enables the sequential autoregressive generation of entities and descriptions from the question until the $\texttt{<eod>}$ token is produced. The overall context, including entities and descriptions, is then considered, and a summary that aligns more closely with the question is generated. This process ultimately constructs the input prompt for the reader model, ensuring a reliable response to the question.

For all datasets, we use a 0-shot setting in our experiments. The prompt examples for the reader model can be found in Table~\ref{tab:reader_prompt}.
\begin{table*}[t]
\small
\resizebox{\textwidth}{!}{%
\begin{tabular}{lcccccccccccc}
\toprule
 & \multicolumn{8}{c}{\textbf{General-purpose LLMs}} & \multicolumn{4}{c}{\textbf{Medical-purpose LLMs}} \\
 & \multicolumn{2}{c}{\textbf{Llama-3-8B}} & \multicolumn{2}{c}{\textbf{Llama-3-70B}} & \multicolumn{2}{c}{\textbf{Mixtral-8x7B}} & \multicolumn{2}{c}{\textbf{GPT-4o}\textsuperscript{~\ref{foot:gpt}}} & \multicolumn{2}{c}{\textbf{MedAlpaca-13B}} & \multicolumn{2}{c}{\textbf{Meditron-70B}} \\
 & \textbf{BertScore} & \textbf{UniEval} & \textbf{BertScore} & \textbf{UniEval} & \textbf{BertScore} & \textbf{UniEval} & \textbf{BertScore} & \textbf{UniEval} & \textbf{BertScore} & \textbf{UniEval} & \textbf{BertScore} & \textbf{UniEval} \\
\midrule
\multicolumn{13}{c}{\underline{MedQuAD}} \\
\multicolumn{13}{l}{\textbf{Without compressor}} \\
Top-1 document & 75.65 & 53.44 & 79.44 & 57.47 & 58.68 & 44.30 & 84.45 & 60.24 & 75.05 & 36.89 & 76.38 & 52.74 \\
Top-5 documents & 76.20 & 52.45 & 79.95 & 55.25 & 64.17 & 47.47 & 83.08 & \textbf{64.98} & 17.57 & 10.82 & 73.14 & 51.88 \\
\midrule
\multicolumn{13}{l}{\textbf{With compressor}} \\
RECOMP & 73.27 & 55.74 & 79.75 & \textbf{63.57} & 71.58 & 58.78 & 85.20 & 60.92 & 82.52 & 39.15 & 76.41 & 56.14 \\
LLMLingua & 76.45 & 54.65 & 78.78 & 57.00 & 59.79 & 45.34 & 84.46 & 55.04 & 83.10 & 44.17 & 74.74 & 54.18 \\
FineTune & 74.27 & 54.89 & 79.97 & 59.67 & 71.44 & 56.26 & 83.47 & 62.40 & 82.25 & 39.40 & 76.14 & 55.97 \\
\rowcolor[HTML]{EFEFEF}
\textsc{K-comp} & \textbf{83.79} & \textbf{64.05} & \textbf{82.06} & 62.91 & \textbf{80.64} & \textbf{63.07} & \textbf{85.49} & 63.16 & \textbf{85.75} & \textbf{49.39} & \textbf{78.88} & \textbf{58.95} \\
\midrule
\multicolumn{13}{c}{\underline{MASH-QA}} \\
\multicolumn{13}{l}{\textbf{Without compressor}} \\
Top-1 document & 77.87 & 54.08 & 82.65 & 56.48 & 72.63 & 51.88 & 83.39 & 60.66 & 76.95 & 43.41 & 80.45 & 55.31 \\
Top-5 documents & 77.59 & 52.94 & 82.07 & 57.20 & 73.09 & 53.18 & 83.12 & \textbf{64.81} & 31.91 & 20.07 & 79.31 & 55.85 \\
\midrule
\multicolumn{13}{l}{\textbf{With compressor}} \\
RECOMP & 80.20 & 58.80 & 82.27 & 60.50 & \textbf{82.12} & 62.35 & 83.82 & 62.22 & 78.51 & 34.83 & 81.54 & 60.09 \\
LLMLingua & 79.74 & 54.37 & 83.25 & 56.83 & 76.71 & 58.32 & 83.37 & 56.73 & 80.41 & 48.45 & 80.78 & 58.55 \\
FineTune & 77.30 & 55.21 & 82.90 & 60.66 & 78.74 & 60.62 & 83.88 & 62.82 & 79.96 & 44.02 & 81.70 & 60.06 \\
\rowcolor[HTML]{EFEFEF}
\textsc{K-comp} & \textbf{83.45} & \textbf{61.45} & \textbf{83.64} & \textbf{61.40} & 81.49 & \textbf{62.91} & \textbf{83.92} & 64.06 & \textbf{81.95} & \textbf{51.76} & \textbf{82.67} & \textbf{61.85} \\
\midrule
\multicolumn{13}{c}{\underline{BioASQ}} \\
\multicolumn{13}{l}{\textbf{Without compressor}} \\
Top-1 document & 82.53 & 56.03 & 81.40 & 59.58 & 82.23 & 62.07 & 87.13 & 62.47 & 76.86 & 40.06 & 83.43 & 57.55 \\
Top-5 documents & 81.44 & 61.33 & 80.91 & 61.09 & 83.54 & 65.53 & 86.81 & \textbf{66.15} & 20.93 & 12.41 & 83.63 & 59.52 \\
\midrule
\multicolumn{13}{l}{\textbf{With compressor}} \\
RECOMP & 83.89 & 59.80 & 85.45 & 59.65 & \textbf{86.85} & 62.82 & 87.87 & 60.62 & 78.07 & 30.68 & 84.77 & 57.58 \\
LLMLingua & 81.70 & 53.62 & 86.53 & 61.53 & 82.54 & 62.24 & 86.85 & 51.11 & 81.19 & 42.97 & 81.51 & 55.26 \\
FineTune & 82.26 & 60.89 & 86.17 & 62.86 & 85.48 & 64.01 & \textbf{87.97} & 63.99 & 78.67 & 41.49 & 85.05 & 61.20 \\
\rowcolor[HTML]{EFEFEF}
\textsc{K-comp} & \textbf{86.61} & \textbf{62.28} & \textbf{87.03} & \textbf{62.89} & 86.52 & \textbf{66.31} & 87.77 & 64.38 & \textbf{84.88} & \textbf{45.47} & \textbf{86.70} & \textbf{62.90} \\
\bottomrule
\end{tabular}%
}
\vspace{-1mm}
\caption{\label{tab:main_result}Main results. We report automatic evaluations for retrieval-augmented QA with and without compressors.}
\vspace{-1mm}
\end{table*}

\section{\label{sec:experiments}Experiments}
In this section, we evaluate \textsc{K-comp} trained by causal knowledge injection and the retrieval-augmented QA task. We report the datasets and settings used in the experiments (\S\ref{sec:Settings}) and discuss the main results (\S\ref{sec:Results}).

\subsection{Settings}
\label{sec:Settings}
\paragraph{Models} We fine-tuned Gemma-2B~\citep{team2024gemma} with our knowledge injection objective. Further details regarding the models and implementation can be found in Appendix~\ref{sec:exp_detail}.
\paragraph{Datasets} To reduce potential biases from fine-tuned medical LLMs~\citep{han2023medalpaca, chen2023meditron}, we conduct experiments using the medical QA datasets MedQuAD~\citep{BenAbacha-BMC-2019}, MASH-QA~\citep{zhu-etal-2020-question}, and BioASQ~\citep{krithara2023bioasq}, which were not directly used for training biomedical models. Although MASH-QA and BioASQ provide gold passages containing answers, our experiments do not utilize these gold passages. Instead, we rely on passages retrieved by our retrieval framework.

\paragraph{Evaluation Metrics} Since all datasets consist of long-form answers, we use the trained model to evaluate the answers. We quantify the relevance of answers using BertScore~\citep{bert-score}, which evaluates the similarity between two sentences by exploiting the contextual embeddings of the encoder. We also use UniEval~\citep{zhong-etal-2022-towards}, which is a multi-dimensional evaluation metric that has high correlation and similarity with human judgment. We explicitly assess the factual consistency between generated and gold answers.

\subsection{Results}
\label{sec:Results}
\paragraph{Baselines} We compare \textsc{K-comp} with standard RAG approach with top-1 and top-5 retrieved passages without applying prompt compression. We also compare with previous state-of-the-art prompt compression methods, including RECOMP~\citep{xu2024recomp} and LLMLingua~\citep{jiang-etal-2023-llmlingua}. Specifically, for implementing RECOMP, we use an abstractive compressor fine-tuned on our datasets, and for LLMLingua, we use Llama-2-7B~\citep{touvron2023llama} for compression. The prompts for synthesizing the summaries used in RECOMP are based on the paper and can be found in Table~\ref{tab:summary_generation_recomp}. Furthermore, the efficacy of causal knowledge injection is evaluated by comparing it to a model that has been fine-tuned (FineTune) using only the standard language modeling objective for summarization. FineTune fine-tuned with Gemma-2B, the same as \textsc{K-comp}.

\paragraph{Overall Performance} Table~\ref{tab:main_result} shows the main results of \textsc{K-comp} compared to the baselines across various reader LLMs. Overall, compression methods are effective. Chunking snippets for retrieval is inherently imperfect, making the Top-1 and Top-5 passages suboptimal. For MedAlpaca, which has the smallest context window size of 2048 among the reader models, the answer accuracy declines significantly with Top-5 passages input due to the limited window size. Consequently, a reprocessing stage, such as compression, is required to improve the quality of chunked text and enable the reader model to reference it appropriately. Among the baselines, LLMLingua lags behind other baselines trained in the medical domain due to its query-agnostic compression approach. We also observe different results depending on the model size. Larger models are less dynamic in their response, relying more on their internal knowledge and less on the prompt variations. As can be seen in the case study (\S\ref{sec:case_study}) and Table~\ref{tab:full_case_study}, larger models are capable of providing reasonable responses to questions even when presented with noisy input prompts. In other words, the result demonstrates that not only small models but also large models place greater trust in the description and concise text provided by \textsc{K-comp} than in other baselines.

To emphasize the importance of entity and description, we analyze a scenario where \textsc{K-comp} infers normally but only appends the summary to the reader prompt (Table~\ref{tab:prior_knowledge}). $-Prior$ is comparable to the baseline fine-tuned for summarization tasks. Even so, it is clear that providing the reader model with prior knowledge significantly improves the accuracy of the final responses compared to FineTune. An interesting observation is that general-purpose models exhibit a more significant influence of information related to medical jargon compared to medical LLMs. Medical LLMs seem to treat knowledge about entities as noisy input, resulting in conflicts with their internal knowledge. Still, these analyses are confined to the QA accuracy of reader LLMs, as they are affected by changes in the components that make up the prompt. The following sections will discuss the relevance and alignment of the summary.
\begin{table*}[t]
\tiny
\renewcommand{\arraystretch}{0.85}
\resizebox{\textwidth}{!}{%
\begin{tabular}{lcccccccc}
\toprule
 & \multicolumn{2}{c}{\textbf{Llama-3-8B}} & \multicolumn{2}{c}{\textbf{Llama-3-70B}} & \multicolumn{2}{c}{\textbf{MedAlpaca-13B}} & \multicolumn{2}{c}{\textbf{Meditron-70B}} \\
 & \textbf{BertScore} & \textbf{UniEval} & \textbf{BertScore} & \textbf{UniEval} & \textbf{BertScore} & \textbf{UniEval} & \textbf{BertScore} & \textbf{UniEval} \\
\midrule
\multicolumn{9}{c}{MedQuAD} \\
\rowcolor[HTML]{EFEFEF}
\textsc{K-comp} & \textbf{83.79} & \textbf{64.05} & \textbf{82.06} & \textbf{62.91} & \textbf{85.75} & \textbf{49.39} & 78.88 & 58.95 \\
$-Prior$ & 73.93 & 54.47 & 80.06 & 60.03 & 84.84 & 48.23 & \textbf{79.80} & \textbf{59.00} \\
FineTune & 74.27 & 54.89 & 79.97 & 59.67 & 82.25 & 39.40 & 76.14 & 55.97 \\
\midrule
\multicolumn{9}{c}{MASH-QA} \\
\rowcolor[HTML]{EFEFEF}
\textsc{K-comp} & \textbf{83.45} & \textbf{61.45} & \textbf{83.64} & \textbf{61.40} & 81.95 & 51.76 & \textbf{82.67} & \textbf{61.85} \\
$-Prior$ & 78.63 & 57.80 & 83.16 & 60.88 & \textbf{82.18} & \textbf{52.02} & 81.93 & 60.42 \\
FineTune & 77.30 & 55.21 & 82.90 & 60.66 & 79.96 & 44.02 & 81.70 & 60.06 \\
\midrule
\multicolumn{9}{c}{BioASQ} \\
\rowcolor[HTML]{EFEFEF}
\textsc{K-comp} & \textbf{86.61} & \textbf{62.28} & \textbf{87.03} & \textbf{62.89} & \textbf{84.88} & 45.47 & \textbf{86.70} & \textbf{62.90} \\
$-Prior$ & 80.75 & 55.06 & 86.22 & 62.40 & 84.53 & \textbf{46.67} & 85.23 & 59.73 \\
FineTune & 82.26 & 60.89 & 86.17 & 62.86 & 78.67 & 41.49 & 85.05 & 61.20 \\
\bottomrule
\end{tabular}%
}
\vspace{-1mm}
\caption{
\label{tab:prior_knowledge} Ablation studies. $-Prior$ denotes the scenario where \textsc{K-comp} does not provide prior knowledge to the reader LLMs.}
\vspace{-1mm}
\end{table*}

\begin{figure}[h]
\small
\centering
\includegraphics[scale=0.35]{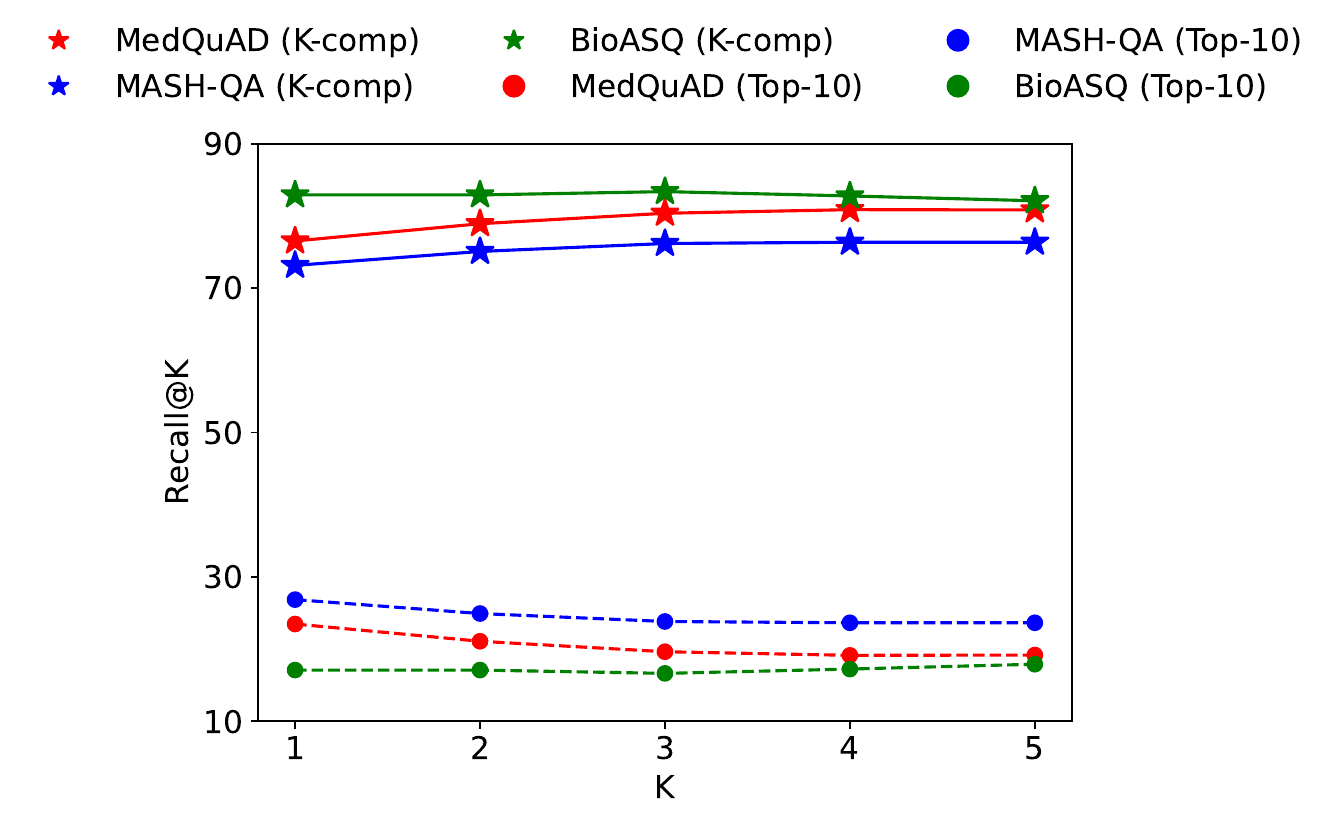}
\vspace{-1mm}
\caption{Percentage of Recall@$K$ according to the variation of $K$ for the retrieved passages and our compressed contexts, where Top-10 denotes the ten retrieved passages with the highest similarity scores.}
\label{fig:rerank}
\vspace{-1mm}
\end{figure}
\section{Analyses}
We analyze the results from various perspectives (\S\ref{sec:reranking}, \ref{sec:speed}, \ref{sec:case_study}). Finally, we appraise the outcomes focusing on the GPT-4o evaluation to highlight the advantages of \textsc{K-comp} through comparison with previous studies (\S\ref{sec:gpt_eval}).

\subsection{Reranking Preference}
\label{sec:reranking}
In addition to evaluating the end-task performance, it is crucial in RAG to ensure that prompts are augmented to be pertinent to the question. Although human evaluation is valuable, it demands significant resources and domain expertise, which are not readily available in our case. Instead, we propose to employ a state-of-the-art reranker\footnote{We use BAAI/bge-reranker-large~\citep{10.1145/3626772.3657878}.} to measure the relevance between the context and the question. For each question $q$, we execute \textsc{K-comp} to generate 10 contexts using a high temperature setting (temperature=1) based on $q_m \oplus \mathcal{P}$. Next, we retrieve the top-10 passages related to $q$. Thus, we gather a total of 20 passages to be fed to the reranker. By applying Recall@$K$ to these 20 passages, we observe the $K$ passages that are most similar to $q$, and quantify the proportion of the compressor varied as $K$ varied.

Figure~\ref{fig:rerank} illustrates Recall@$K$ across different values of $K$. Specifically, we achieved Recall@1 scores of 77\%, 73\%, and 83\% on MedQuAD, MASH-QA, and BioASQ, whereas the top-5 retrieved passages achieved 23\%, 27\%, and 17\%. This comparison demonstrates that the reranker strongly prefers our compressed contexts across all three benchmarks.

\begin{table}[t]
\small
\centering
\renewcommand{\arraystretch}{0.85}
\resizebox{\columnwidth}{!}{%
\begin{tabular}{l|ccc}
\toprule
\textbf{Settings} & \textbf{Top-1} & \textbf{Top-5} & \textbf{\textsc{K-comp}} \\
\midrule
Input tokens & 195 & 978 & 198 \\
Inference time & 9,798s & 20,193s & 9,777s \\
Compression time & - & - & 497s \\
Total time & 9,798s & 20,193s & 10,274s \\
\bottomrule
\end{tabular}}
\vspace{-1mm}
\caption{Inference speed of Llama-3-70B on MASH-QA.}
\label{tab:inference_speed}
\vspace{-1mm}
\end{table}
\begin{table*}[t]
\centering
\small
\renewcommand{\arraystretch}{0.85}
\resizebox{\textwidth}{!}{%
\begin{tabular}{l|p{6cm}|p{6cm}}
\toprule
\multirow{2}{*}{} & \multicolumn{2}{c}{Question: How do \textbf{\textcolor{red}{x-rays}} diagnose \textbf{\textcolor{red}{rheumatoid arthritis}} (RA)?} \\ \cmidrule{2-3}
 & \multicolumn{1}{l|}{Augmented Prompt} & {Answer} \\
\midrule
FineTune & \multicolumn{1}{p{6cm}|}{\#\#\# Passage\newline Rheumatoid arthritis (RA) is a chronic systemic disease characterized by inflammation of the synovial tissue, affecting 0.5\%-1.0\% of the global population, with ...} & X-rays are used to diagnose RA by detecting joint damage and erosion. \\
\midrule
\textsc{K-comp} & \multicolumn{1}{p{6cm}|}{\#\#\# Entity\newline \textbf{\textcolor{red}{x-ray}}:\{\{description\}\}\newline \textbf{\textcolor{red}{rheumatoid arthritis}}: \{\{description\}\}\newline\newline \#\#\# Passage\newline \textbf{\textcolor{red}{X-rays}} are the most common method for assessing the degree of joint destruction in \textbf{\textcolor{red}{rheumatoid arthritis (RA)}}, revealing osteopenia and joint space narrowing as the disease ...} & X-rays diagnose rheumatoid arthritis (RA) by revealing changes in the axial skeleton and joint space. As the disease progresses, early RA may not show significant changes on X-rays, but it can indicate osteopenia and joint space narrowing. Advances in imaging techniques like MRI and ultrasound enhance the ability to detect soft tissue changes and joint damage, which is crucial for accurate diagnosis ...\\
\bottomrule
\end{tabular}%
}
\vspace{-1mm}
\caption{Case study. We provide the passages used to augment the reader's prompt and the answers. \textbf{\textcolor{red}{Red texts}} highlight the medical jargon within the question. The complete prompt can be found in Table~\ref{tab:full_case_study}.}
\label{tab:case}
\vspace{-1mm}
\end{table*}

\subsection{Inference Speed}
\label{sec:speed}
In Table~\ref{tab:inference_speed}, we report the inference time and the number of tokens used in the prompt input as metrics for evaluating efficiency. Specifically, we employed Llama-3-70B as the reader model and measured the GPU runtime on MASH-QA test set. Both the compressor and reader are executed on a single NVIDIA A100 GPU with 80GB memory. Even when considering the time needed for the compressor inference, our method was able to double the throughput compared to prepending the top-5 passages, making it more efficient. Moreover, we note that inference speed is dependent on the implementation and size of the reader model. For instance, models with more parameters will suffer increased latency by increasing the number of input tokens. This phenomenon amplifies the speed advantage of \textsc{K-comp}.

\subsection{Case Study}
\label{sec:case_study}
Here, we report how \textsc{K-comp} generates medical knowledge. In Table~\ref{tab:case}, \textsc{K-comp} is able to address the practical utilization of X-rays in the diagnosis and monitoring of RA, and provides detailed explanations regarding their application, which offers a more transparent rationale for addressing the questions. By contrast, FineTune provides a more general context without focusing on the specific role of X-rays in RA diagnosis. Although it mentions X-rays along with other diagnostic techniques, its focus is on advancements in imaging methods such as MRI. FineTune merely summarises the passages retrieved based on semantic and overall lexical similarities to the question without considering the queried intent. This results in the reader model does not fully trusting the augmented passages, instead perceiving them as irrelevant noise and generating answers not based on the passages. This result can lead to inaccuracies and potential hallucinations.

\subsection{GPT-4o Evaluation}
\label{sec:gpt_eval}
We additionally explore the reliability of the context. Given that GPT-4 has been demonstrated to correlate highly with human judgments~\citep{liu-etal-2023-g}, even in the medical domain~\citep{nori2023capabilitiesgpt4medicalchallenge}, we employed GPT-4o\footnote{\label{foot:gpt}We use gpt-4o-2024-05-13~\citep{hurst2024gpt}.} to perform a comparative evaluation of summaries generated by baselines and \textsc{K-comp}. All prompts utilized in the evaluation were structured using the identical format as presented in Table~\ref{tab:gpt_eval}. Finally, we report examples of results for all baselines in Table~\ref{tab:example_baseline} for qualitative analysis.

\begin{figure*}[t]
\centering
\includegraphics[width=\linewidth]{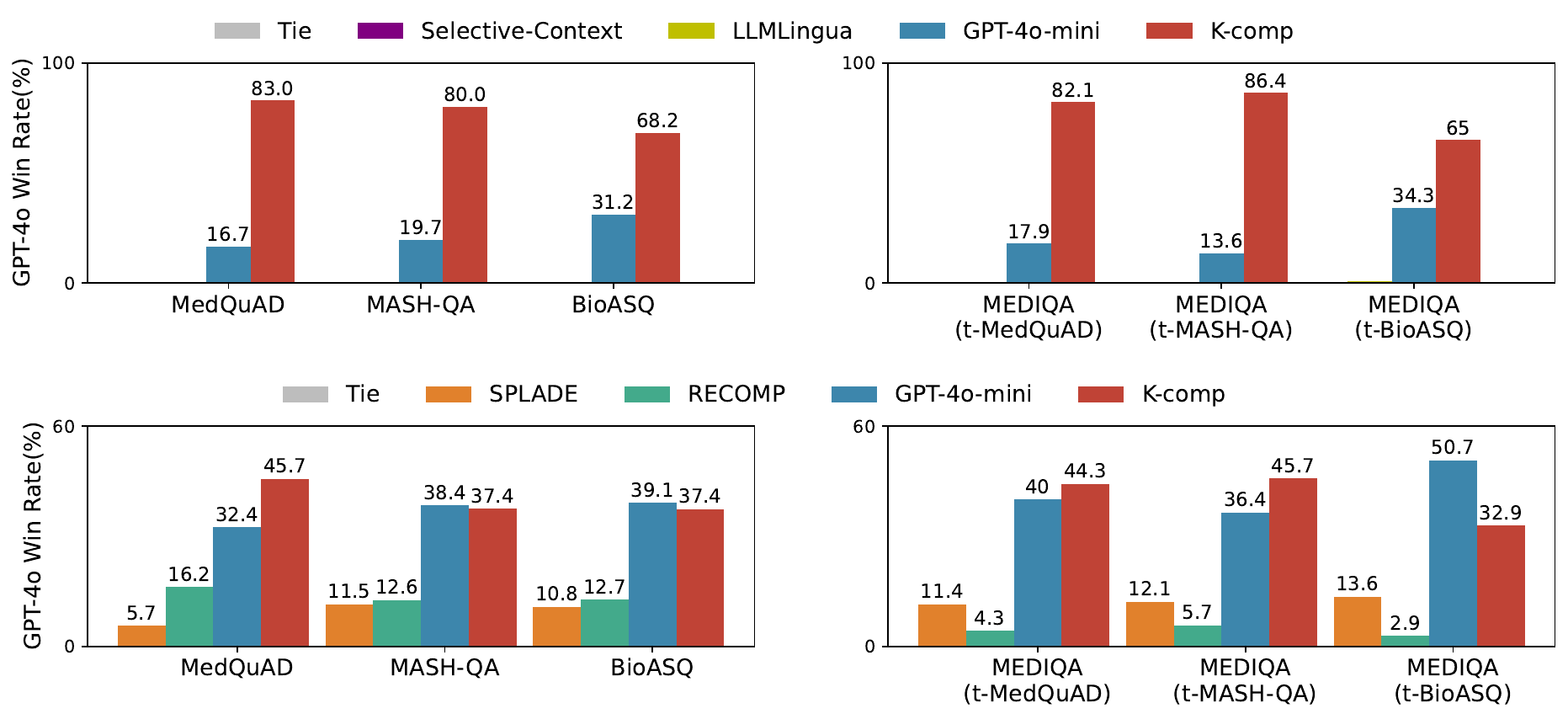} 
\vspace{-1mm}
\caption{GPT-4o evaluation results with baselines. t-BioASQ indicates that MEDIQA was inferred using \textsc{K-comp} (or RECOMP) trained on BioASQ. For clarity, results in the 0\% range are not indicated. Accordingly, we report all results in Table~\ref{tab:full_gpt_eval}. (Above) Query-agnostic prompt compression, (Below) Query-based prompt compression methods, (Left) Results from the seen data, (Right) Results from the unseen data.}
\label{fig:gpt_eval_baseline}
\vspace{-1mm}
\end{figure*}

\paragraph{Query-Agnostic}
Figure~\ref{fig:gpt_eval_baseline} compares our approach to previous studies that compress prompts both with and without relying on the query. Intuitively, the baselines compared in query-agnostic section are compressed regardless of the question, which demonstrates that \textsc{K-comp} outperforms the other methods because it references the masked question. Detailed analyses are provided in Appendix~\ref{sec:query-agnostic}.

\paragraph{Query-Based}
SPLADE~\citep{10.1145/3477495.3531833} is a lexical-based retriever that reweights each term by emphasizing important terms associated with the query. The most informative top-1 passages among the top-5 retrieved passages are extracted and assumed to be compressed context. GPT-4o-mini\textsuperscript{~\ref{foot:gpt-mini}} is the result of a summary generated using prompt~\ref{tab:summary_generation_recomp}, which is the same prompt used when synthesizing the RECOMP train data. The rationale for \textsc{K-comp} being rated higher than RECOMP is that even when trained on that dataset, it produces a summary that lacks specialization, which is similar to the output of FineTune in case study. By using a seq2seq model rather than token-level pruning such as LLMLingua, it allows the delivery of complete sentences to the reader LLM in QA tasks. This prevents the LLM from perceiving excessive noise, thereby ensuring relatively strong end-task performance among the baselines (Table~\ref{tab:main_result}). However, due to the maximum token limitation, the generated text is shorter and contains less information in the context, which results in lower evaluation scores for summary quality.

\textsc{K-comp} demonstrates comparable performance to GPT-4o-mini, particularly exhibiting significant superiority on MedQuAD. In query-based comparisons, GPT-4o-mini generates a rationale that is highly effective for answering based on its demonstrated capabilities in text generation. Similarly, \textsc{K-comp} exhibits performance comparable to that of GPT-4o-mini, despite being composed of only 2B parameters.

\paragraph{Unseen Evaluation}
In order to provide additional novelty to our approach, we evaluate baselines on data that was not used during training. Unlike query-based methods, such as RECOMP, which are trained to generate summaries optimized for responding to questions, our approach, based on the entities, demonstrates efficacy when applied to unseen data. The right-hand column of Figure~\ref{fig:gpt_eval_baseline} illustrates the results of applying each baseline to 140 test data from the MEDIQA~\citep{ben-abacha-etal-2019-overview}. As discussed in Appendix~\ref{sec:query-agnostic}, Selective-Context~\citep{li-etal-2023-compressing} and LLMLingua generate incomplete sentences, which result in inferior performance despite their compression in a question-agnostic fashion. In contrast, \textsc{K-comp} maintains a competitive performance on unseen data, comparable to the results evaluated on seen data.

A noteworthy point is the comparison result with query-based baselines. As can be seen in Table~\ref{tab:summary_generation} and~\ref{tab:summary_generation_recomp}, the instructions used for synthesizing summaries in RECOMP and \textsc{K-comp} were as follows: "Compress ... used to answer the question" and "Extract the content about the entity", respectively. By focusing on the entities, the objective of our training approach is to provide a concise context of the medical terminology that has been requested in the question. As a consequence, \textsc{K-comp} achieves a win rate that is similar to, and even exceeds, the results obtained in the training datasets. This presents a limitation of previous methods, which are unable to maintain their performance levels due to their reliance on the distribution of training data. On the other side, our training approach focuses on medical jargon, which is advantageous because the medical terminology remains consistent even when the data changes. Therefore, our causal knowledge injection substantially contributes to improving performance in data-scarce, closed-domain settings.

\paragraph{Additional NLG Evaluation}
\begin{table}[t]
\renewcommand{\arraystretch}{0.85}
\resizebox{\columnwidth}{!}{%
\begin{tabular}{lcccccc}
\toprule
 & \textbf{Llama-8B} & \textbf{Llama-70B} & \textbf{Mixtral} & \textbf{GPT-4o} & \textbf{MedAlpaca} & \textbf{Meditron} \\ \midrule
\multicolumn{7}{c}{\underline{MedQuAD}} \\
\multicolumn{7}{l}{\textbf{Without compressor}} \\
Top-1 doc & 2.96 & 3.11 & 2.63 & 4.27 & 1.65 & 3.14 \\
Top-5 docs & 3.22 & 3.36 & 3.07 & \textbf{4.73} & 1.85 & 3.53 \\ \midrule
\multicolumn{7}{l}{\textbf{With compressor}} \\
RECOMP & 3.29 & \textbf{3.80} & 3.51 & 3.92 & 1.48 & 3.68 \\
LLMLingua & 2.41 & 2.87 & 2.53 & 3.68 & 1.61 & 3.21 \\
FineTune & 3.28 & 3.66 & 3.27 & 4.15 & 1.44 & 3.68 \\
\rowcolor[HTML]{EFEFEF}
\textsc{K-comp} & \textbf{3.55} & 3.66 & \textbf{3.73} & 4.48 & \textbf{1.98} & \textbf{3.85} \\ \midrule
\multicolumn{7}{c}{\underline{MASH-QA}} \\
\multicolumn{7}{l}{\textbf{Without compressor}} \\
Top-1 doc & 3.07 & 3.65 & 3.24 & 4.63 & 1.64 & 3.55 \\
Top-5 docs & 3.33 & 3.58 & 3.46 & \textbf{4.74} & 1.79 & 3.43 \\ \midrule
\multicolumn{7}{l}{\textbf{With compressor}} \\
RECOMP & 3.68 & 4.14 & \textbf{4.49} & 4.15 & 1.34 & 4.01 \\
LLMLingua & 2.97 & 3.18 & 3.77 & 3.99 & 1.88 & 3.47 \\
FineTune & 3.36 & 3.79 & 3.78 & 3.93 & 1.74 & 3.51 \\
\rowcolor[HTML]{EFEFEF}
\textsc{K-comp} & \textbf{4.23} & \textbf{4.62} & 4.06 & 4.52 & \textbf{1.99} & \textbf{4.53} \\ \midrule
\multicolumn{7}{c}{\underline{BioASQ}} \\
\multicolumn{7}{l}{\textbf{Without compressor}} \\
Top-1 doc & 3.44 & 3.75 & 3.64 & 4.33 & 2.09 & 3.76 \\
Top-5 docs & 3.58 & 3.66 & 3.68 & \textbf{4.53} & 1.32 & 3.64 \\ \midrule
\multicolumn{7}{l}{\textbf{With compressor}} \\
RECOMP & 3.64 & 3.92 & \textbf{4.10} & 3.58 & 1.18 & 4.00 \\
LLMLingua & 2.54 & 3.63 & 3.54 & 3.37 & 1.87 & 3.42 \\
FineTune & 3.63 & 4.05 & 3.59 & 4.01 & 1.48 & 3.88 \\
\rowcolor[HTML]{EFEFEF}
\textsc{K-comp} & \textbf{4.10} & \textbf{4.14} & 3.81 & 4.20 & \textbf{2.31} & \textbf{4.38} \\ \bottomrule
\end{tabular}%
}
\vspace{-1mm}
\caption{\label{tab:gpt_eval_table} Results evaluated using the G-Eval-4 metric~\citep{liu-etal-2023-g}. The prompt used is presented in Table~\ref{tab:g_eval}.}
\vspace{-1mm}
\end{table}
The validity of our method has been established through rigorous empirical validation employing the BertScore and UniEval evaluation metrics. To further enhance methodological consistency and ensure a comprehensive evaluation, we have incorporated G-Eval~\citep{liu-etal-2023-g}, a cutting-edge assessment metric, to perform an extensive supplementary analysis. This evaluation is conducted leveraging a highly advanced commercial large language model (LLM), thereby reinforcing the reliability and validity of the proposed approach. For G-Eval-4, GPT-4 is sampled 20 times, with the resulting average score used to minimize the impact of any potential variability. Given the broad scope of our methodological evaluation, which covers diverse datasets and a wide range of models, evaluating a full-scale analysis with G-Eval would be extremely computationally expensive. As a result, 1k data points per dataset were randomly selected as the test data. The results are reported in Table~\ref{tab:gpt_eval_table}. Regarding performance, \textsc{K-comp} consistently exhibits superior performance compared to other baselines.

Interestingly, our results align closely with the trends observed in UniEval. According to the G-Eval study, UniEval exhibited a stronger correlation with human judgments in summary, dialogue generation, and consistency evaluation than all baselines except G-Eval-4. This trend is also reflected in our findings. In Table~\ref{tab:main_result}, when GPT-4o is used as the reader model, the responses generated by augmenting the top-5 documents achieve the highest scores, a pattern consistent with G-Eval. Additionally, on the MedQuAD dataset, when Llama-3-70B serves as the reader, UniEval shows a preference for RECOMP, which aligns with the corresponding G-Eval results. Similarly, in other scenarios, both metrics indicate a preference for \textsc{K-comp}. These observations suggest that UniEval effectively mirrors G-Eval despite being a relatively older metric. While our study does not include human evaluation, the strong alignment between UniEval and G-Eval suggests that our methodology is likely to correlate well with human judgments.
\section{Conclusion} In this paper, we have proposed a novel method to improve retrieval-augmented QA by compressing retrieved documents focused on the questions. We have devised a comprehensive scheme for identifying medical entities and automatically generating prior knowledge. This is followed by the extension of training and inference methods, which enable the autoregressive generation of summaries that incorporate domain knowledge while considering the context causally. Furthermore, we proved that this approach is practical even when applied to unseen evaluation, which represents a novel contribution in closed domains where data is scarce.
\section*{Acknowledgements}
This work was supported by Institute of Information \& communications Technology Planning \& Evaluation (IITP) grant funded by the Korea government(MSIT) (No.RS-2019-II191906, Artificial Intelligence Graduate School Program(POSTECH))

This research was supported by Smart HealthCare for Police Officers Program(www.kipot.or.kr) through the Korea Institutes of Police Technology(KIPoT) funded by the Korean National Police Agency(KNPA, Korea)(No. RS-2022-PT000186)

\section*{Limitations}
Our methodology is limited in scenarios where the NER tool is unable to automatically detect ambiguous keywords or entities that are absent from the questions. To mitigate these issues, expanding the retrieval corpus with additional text chunks can inject more knowledge into the compressor and learn domain-relevant entities, but this will drastically increase the cost of annotating the data and require enormous resources for retrieval to perform nearest-neighbor searches. Therefore, we consider extending these retrieval datastores an important task in RAG, and this can be extended in future work.

Additionally, our study mainly focuses on English biomedical QA, which limits generalization to other languages and domains. Current studies in closed domains face challenges due to the scarcity of datasets, posing a considerable obstacle to the broader implementation of our methodology. We believe that, among closed domains, the medical QA has relatively more data, and we have proven our methodology in this domain. However, in other specific domains, not only QA data but also retrieval corpora are yet to be established. Furthermore, data availability in languages other than English is even more limited. Nevertheless, we recognize that our methodology has significant potential for extension to other languages and domains and that such expansion is necessary to demonstrate the generalizability of our training approach. Accordingly, we regard the application of retrieval-augmented QA in closed domains as a critical area of investigation, so we intend to extend our research to encompass additional domains in the future.

\section*{Ethical Considerations} In our research, we employed publicly available datasets, including MedQuAD, MASH-QA, BioASQ, and MEDIQA. When synthesizing ground-truth summaries, we ensure that no personally identifiable information is used and that all data are anonymized. Our methodology is still in its early stages and is not yet suitable for direct practical use in medical domains, where reliability and accuracy are paramount. In particular, hallucination can critically affect patient care and clinical decision-making. Therefore, our methodology is considered to mitigate hallucination by emphasizing domain knowledge in biomedical QA research rather than substituting professional medical judgment, thus posing no risk of harm.
\nocite{*}
\bibliography{custom}

\begin{thebibliography}{79}
\providecommand{\natexlab}[1]{#1}

\bibitem[{Aghajanyan et~al.(2022)Aghajanyan, Huang, Ross, Karpukhin, Xu, Goyal, Okhonko, Joshi, Ghosh, Lewis, and Zettlemoyer}]{aghajanyan2022cm3causalmaskedmultimodal}
Armen Aghajanyan, Bernie Huang, Candace Ross, Vladimir Karpukhin, Hu~Xu, Naman Goyal, Dmytro Okhonko, Mandar Joshi, Gargi Ghosh, Mike Lewis, and Luke Zettlemoyer. 2022.
\newblock \href {https://arxiv.org/abs/2201.07520} {Cm3: A causal masked multimodal model of the internet}.
\newblock \emph{Preprint}, arXiv:2201.07520.

\bibitem[{Ahmad et~al.(2019)Ahmad, Constant, Yang, and Cer}]{ahmad-etal-2019-reqa}
Amin Ahmad, Noah Constant, Yinfei Yang, and Daniel Cer. 2019.
\newblock \href {https://doi.org/10.18653/v1/D19-5819} {{R}e{QA}: An evaluation for end-to-end answer retrieval models}.
\newblock In \emph{Proceedings of the 2nd Workshop on Machine Reading for Question Answering}, pages 137--146, Hong Kong, China. Association for Computational Linguistics.

\bibitem[{AI@Meta(2024)}]{llama3modelcard}
AI@Meta. 2024.
\newblock \href {https://github.com/meta-llama/llama3/blob/main/MODEL_CARD.md} {Llama 3 model card}.

\bibitem[{Asai et~al.(2024{\natexlab{a}})Asai, Wu, Wang, Sil, and Hajishirzi}]{selfrag}
Akari Asai, Zeqiu Wu, Yizhong Wang, Avirup Sil, and Hannaneh Hajishirzi. 2024{\natexlab{a}}.
\newblock \href {https://openreview.net/forum?id=hSyW5go0v8} {Self-{RAG}: Learning to retrieve, generate, and critique through self-reflection}.
\newblock In \emph{The Twelfth International Conference on Learning Representations}.

\bibitem[{Asai et~al.(2024{\natexlab{b}})Asai, Zhong, Chen, Koh, Zettlemoyer, Hajishirzi, and tau Yih}]{asai2024reliableadaptableattributablelanguage}
Akari Asai, Zexuan Zhong, Danqi Chen, Pang~Wei Koh, Luke Zettlemoyer, Hannaneh Hajishirzi, and Wen tau Yih. 2024{\natexlab{b}}.
\newblock \href {https://arxiv.org/abs/2403.03187} {Reliable, adaptable, and attributable language models with retrieval}.
\newblock \emph{Preprint}, arXiv:2403.03187.

\bibitem[{Bavarian et~al.(2022)Bavarian, Jun, Tezak, Schulman, McLeavey, Tworek, and Chen}]{bavarian2022efficienttraininglanguagemodels}
Mohammad Bavarian, Heewoo Jun, Nikolas Tezak, John Schulman, Christine McLeavey, Jerry Tworek, and Mark Chen. 2022.
\newblock \href {https://arxiv.org/abs/2207.14255} {Efficient training of language models to fill in the middle}.
\newblock \emph{Preprint}, arXiv:2207.14255.

\bibitem[{{Ben Abacha} and Demner{-}Fushman(2019)}]{BenAbacha-BMC-2019}
Asma {Ben Abacha} and Dina Demner{-}Fushman. 2019.
\newblock \href {https://bmcbioinformatics.biomedcentral.com/articles/10.1186/s12859-019-3119-4} {A question-entailment approach to question answering}.
\newblock \emph{{BMC} Bioinform.}, 20(1):511:1--511:23.

\bibitem[{Ben~Abacha et~al.(2019)Ben~Abacha, Shivade, and Demner-Fushman}]{ben-abacha-etal-2019-overview}
Asma Ben~Abacha, Chaitanya Shivade, and Dina Demner-Fushman. 2019.
\newblock \href {https://doi.org/10.18653/v1/W19-5039} {Overview of the {MEDIQA} 2019 shared task on textual inference, question entailment and question answering}.
\newblock In \emph{Proceedings of the 18th BioNLP Workshop and Shared Task}, pages 370--379, Florence, Italy. Association for Computational Linguistics.

\bibitem[{Brown et~al.(2020)Brown, Mann, Ryder, Subbiah, Kaplan, Dhariwal, Neelakantan, Shyam, Sastry, Askell, Agarwal, Herbert-Voss, Krueger, Henighan, Child, Ramesh, Ziegler, Wu, Winter, Hesse, Chen, Sigler, Litwin, Gray, Chess, Clark, Berner, McCandlish, Radford, Sutskever, and Amodei}]{NEURIPS2020_1457c0d6}
Tom Brown, Benjamin Mann, Nick Ryder, Melanie Subbiah, Jared~D Kaplan, Prafulla Dhariwal, Arvind Neelakantan, Pranav Shyam, Girish Sastry, Amanda Askell, Sandhini Agarwal, Ariel Herbert-Voss, Gretchen Krueger, Tom Henighan, Rewon Child, Aditya Ramesh, Daniel Ziegler, Jeffrey Wu, Clemens Winter, Chris Hesse, Mark Chen, Eric Sigler, Mateusz Litwin, Scott Gray, Benjamin Chess, Jack Clark, Christopher Berner, Sam McCandlish, Alec Radford, Ilya Sutskever, and Dario Amodei. 2020.
\newblock \href {https://proceedings.neurips.cc/paper_files/paper/2020/file/1457c0d6bfcb4967418bfb8ac142f64a-Paper.pdf} {Language models are few-shot learners}.
\newblock In \emph{Advances in Neural Information Processing Systems}, volume~33, pages 1877--1901. Curran Associates, Inc.

\bibitem[{Chen et~al.(2023)Chen, Cano, Romanou, Bonnet, Matoba, Salvi, Pagliardini, Fan, K{\"o}pf, Mohtashami et~al.}]{chen2023meditron}
Zeming Chen, Alejandro~Hern{\'a}ndez Cano, Angelika Romanou, Antoine Bonnet, Kyle Matoba, Francesco Salvi, Matteo Pagliardini, Simin Fan, Andreas K{\"o}pf, Amirkeivan Mohtashami, et~al. 2023.
\newblock Meditron-70b: Scaling medical pretraining for large language models.
\newblock \emph{arXiv preprint arXiv:2311.16079}.

\bibitem[{Devlin et~al.(2019)Devlin, Chang, Lee, and Toutanova}]{devlin-etal-2019-bert}
Jacob Devlin, Ming-Wei Chang, Kenton Lee, and Kristina Toutanova. 2019.
\newblock \href {https://doi.org/10.18653/v1/N19-1423} {{BERT}: Pre-training of deep bidirectional transformers for language understanding}.
\newblock In \emph{Proceedings of the 2019 Conference of the North {A}merican Chapter of the Association for Computational Linguistics: Human Language Technologies, Volume 1 (Long and Short Papers)}, pages 4171--4186, Minneapolis, Minnesota. Association for Computational Linguistics.

\bibitem[{Donahue et~al.(2020)Donahue, Lee, and Liang}]{donahue-etal-2020-enabling}
Chris Donahue, Mina Lee, and Percy Liang. 2020.
\newblock \href {https://doi.org/10.18653/v1/2020.acl-main.225} {Enabling language models to fill in the blanks}.
\newblock In \emph{Proceedings of the 58th Annual Meeting of the Association for Computational Linguistics}, pages 2492--2501, Online. Association for Computational Linguistics.

\bibitem[{Du et~al.(2022)Du, Qian, Liu, Ding, Qiu, Yang, and Tang}]{du-etal-2022-glm}
Zhengxiao Du, Yujie Qian, Xiao Liu, Ming Ding, Jiezhong Qiu, Zhilin Yang, and Jie Tang. 2022.
\newblock \href {https://doi.org/10.18653/v1/2022.acl-long.26} {{GLM}: General language model pretraining with autoregressive blank infilling}.
\newblock In \emph{Proceedings of the 60th Annual Meeting of the Association for Computational Linguistics (Volume 1: Long Papers)}, pages 320--335, Dublin, Ireland. Association for Computational Linguistics.

\bibitem[{Fried et~al.(2023)Fried, Aghajanyan, Lin, Wang, Wallace, Shi, Zhong, Yih, Zettlemoyer, and Lewis}]{fried2023incoder}
Daniel Fried, Armen Aghajanyan, Jessy Lin, Sida Wang, Eric Wallace, Freda Shi, Ruiqi Zhong, Scott Yih, Luke Zettlemoyer, and Mike Lewis. 2023.
\newblock \href {https://openreview.net/forum?id=hQwb-lbM6EL} {Incoder: A generative model for code infilling and synthesis}.
\newblock In \emph{The Eleventh International Conference on Learning Representations}.

\bibitem[{Frisoni et~al.(2024)Frisoni, Cocchieri, Presepi, Moro, and Meng}]{frisoni-etal-2024-generate}
Giacomo Frisoni, Alessio Cocchieri, Alex Presepi, Gianluca Moro, and Zaiqiao Meng. 2024.
\newblock \href {https://aclanthology.org/2024.acl-long.533} {To generate or to retrieve? on the effectiveness of artificial contexts for medical open-domain question answering}.
\newblock In \emph{Proceedings of the 62nd Annual Meeting of the Association for Computational Linguistics (Volume 1: Long Papers)}, pages 9878--9919, Bangkok, Thailand. Association for Computational Linguistics.

\bibitem[{Guo et~al.(2021)Guo, Yang, Cer, Shen, and Constant}]{guo-etal-2021-multireqa}
Mandy Guo, Yinfei Yang, Daniel Cer, Qinlan Shen, and Noah Constant. 2021.
\newblock \href {https://aclanthology.org/2021.adaptnlp-1.10} {{M}ulti{R}e{QA}: A cross-domain evaluation for{R}etrieval question answering models}.
\newblock In \emph{Proceedings of the Second Workshop on Domain Adaptation for NLP}, pages 94--104, Kyiv, Ukraine. Association for Computational Linguistics.

\bibitem[{Han et~al.(2023)Han, Adams, Papaioannou, Grundmann, Oberhauser, L{\"o}ser, Truhn, and Bressem}]{han2023medalpaca}
Tianyu Han, Lisa~C Adams, Jens-Michalis Papaioannou, Paul Grundmann, Tom Oberhauser, Alexander L{\"o}ser, Daniel Truhn, and Keno~K Bressem. 2023.
\newblock Medalpaca--an open-source collection of medical conversational ai models and training data.
\newblock \emph{arXiv preprint arXiv:2304.08247}.

\bibitem[{Holtzman et~al.(2020)Holtzman, Buys, Du, Forbes, and Choi}]{Holtzman2020The}
Ari Holtzman, Jan Buys, Li~Du, Maxwell Forbes, and Yejin Choi. 2020.
\newblock \href {https://openreview.net/forum?id=rygGQyrFvH} {The curious case of neural text degeneration}.
\newblock In \emph{International Conference on Learning Representations}.

\bibitem[{Hu et~al.(2024)Hu, Shen, Zhang, Chen, and Tao}]{hu2024transforming}
Shengchao Hu, Li~Shen, Ya~Zhang, Yixin Chen, and Dacheng Tao. 2024.
\newblock On transforming reinforcement learning with transformers: The development trajectory.
\newblock \emph{IEEE Transactions on Pattern Analysis and Machine Intelligence}.

\bibitem[{Hurst et~al.(2024)Hurst, Lerer, Goucher, Perelman, Ramesh, Clark, Ostrow, Welihinda, Hayes, Radford et~al.}]{hurst2024gpt}
Aaron Hurst, Adam Lerer, Adam~P Goucher, Adam Perelman, Aditya Ramesh, Aidan Clark, AJ~Ostrow, Akila Welihinda, Alan Hayes, Alec Radford, et~al. 2024.
\newblock Gpt-4o system card.
\newblock \emph{arXiv preprint arXiv:2410.21276}.

\bibitem[{Izacard et~al.(2022)Izacard, Caron, Hosseini, Riedel, Bojanowski, Joulin, and Grave}]{izacard2022unsuperviseddenseinformationretrieval}
Gautier Izacard, Mathilde Caron, Lucas Hosseini, Sebastian Riedel, Piotr Bojanowski, Armand Joulin, and Edouard Grave. 2022.
\newblock \href {https://arxiv.org/abs/2112.09118} {Unsupervised dense information retrieval with contrastive learning}.
\newblock \emph{Preprint}, arXiv:2112.09118.

\bibitem[{Ji et~al.(2023{\natexlab{a}})Ji, Lee, Frieske, Yu, Su, Xu, Ishii, Bang, Madotto, and Fung}]{10.1145/3571730}
Ziwei Ji, Nayeon Lee, Rita Frieske, Tiezheng Yu, Dan Su, Yan Xu, Etsuko Ishii, Ye~Jin Bang, Andrea Madotto, and Pascale Fung. 2023{\natexlab{a}}.
\newblock \href {https://doi.org/10.1145/3571730} {Survey of hallucination in natural language generation}.
\newblock \emph{ACM Comput. Surv.}, 55(12).

\bibitem[{Ji et~al.(2023{\natexlab{b}})Ji, Yu, Xu, Lee, Ishii, and Fung}]{ji-etal-2023-towards}
Ziwei Ji, Tiezheng Yu, Yan Xu, Nayeon Lee, Etsuko Ishii, and Pascale Fung. 2023{\natexlab{b}}.
\newblock \href {https://doi.org/10.18653/v1/2023.findings-emnlp.123} {Towards mitigating {LLM} hallucination via self reflection}.
\newblock In \emph{Findings of the Association for Computational Linguistics: EMNLP 2023}, pages 1827--1843, Singapore. Association for Computational Linguistics.

\bibitem[{Jiang et~al.(2024)Jiang, Sablayrolles, Roux, Mensch, Savary, Bamford, Chaplot, Casas, Hanna, Bressand et~al.}]{jiang2024mixtral}
Albert~Q Jiang, Alexandre Sablayrolles, Antoine Roux, Arthur Mensch, Blanche Savary, Chris Bamford, Devendra~Singh Chaplot, Diego de~las Casas, Emma~Bou Hanna, Florian Bressand, et~al. 2024.
\newblock Mixtral of experts.
\newblock \emph{arXiv preprint arXiv:2401.04088}.

\bibitem[{Jiang et~al.(2023{\natexlab{a}})Jiang, Wu, Lin, Yang, and Qiu}]{jiang-etal-2023-llmlingua}
Huiqiang Jiang, Qianhui Wu, Chin-Yew Lin, Yuqing Yang, and Lili Qiu. 2023{\natexlab{a}}.
\newblock \href {https://doi.org/10.18653/v1/2023.emnlp-main.825} {{LLML}ingua: Compressing prompts for accelerated inference of large language models}.
\newblock In \emph{Proceedings of the 2023 Conference on Empirical Methods in Natural Language Processing}, pages 13358--13376, Singapore. Association for Computational Linguistics.

\bibitem[{Jiang et~al.(2023{\natexlab{b}})Jiang, Xu, Gao, Sun, Liu, Dwivedi-Yu, Yang, Callan, and Neubig}]{jiang-etal-2023-active}
Zhengbao Jiang, Frank Xu, Luyu Gao, Zhiqing Sun, Qian Liu, Jane Dwivedi-Yu, Yiming Yang, Jamie Callan, and Graham Neubig. 2023{\natexlab{b}}.
\newblock \href {https://doi.org/10.18653/v1/2023.emnlp-main.495} {Active retrieval augmented generation}.
\newblock In \emph{Proceedings of the 2023 Conference on Empirical Methods in Natural Language Processing}, pages 7969--7992, Singapore. Association for Computational Linguistics.

\bibitem[{Jin et~al.(2021)Jin, Pan, Oufattole, Weng, Fang, and Szolovits}]{app11146421}
Di~Jin, Eileen Pan, Nassim Oufattole, Wei-Hung Weng, Hanyi Fang, and Peter Szolovits. 2021.
\newblock \href {https://doi.org/10.3390/app11146421} {What disease does this patient have? a large-scale open domain question answering dataset from medical exams}.
\newblock \emph{Applied Sciences}, 11(14).

\bibitem[{Jin et~al.(2024)Jin, Cao, Chen, Liu, Jiang, Xu, Qiuxia, and Zhao}]{jin-etal-2024-tug}
Zhuoran Jin, Pengfei Cao, Yubo Chen, Kang Liu, Xiaojian Jiang, Jiexin Xu, Li~Qiuxia, and Jun Zhao. 2024.
\newblock \href {https://aclanthology.org/2024.lrec-main.1466} {Tug-of-war between knowledge: Exploring and resolving knowledge conflicts in retrieval-augmented language models}.
\newblock In \emph{Proceedings of the 2024 Joint International Conference on Computational Linguistics, Language Resources and Evaluation (LREC-COLING 2024)}, pages 16867--16878, Torino, Italia. ELRA and ICCL.

\bibitem[{Johnson et~al.(2019)Johnson, Douze, and J{\'e}gou}]{johnson2019billion}
Jeff Johnson, Matthijs Douze, and Herv{\'e} J{\'e}gou. 2019.
\newblock Billion-scale similarity search with {GPUs}.
\newblock \emph{IEEE Transactions on Big Data}, 7(3):535--547.

\bibitem[{Joshi et~al.(2020)Joshi, Chen, Liu, Weld, Zettlemoyer, and Levy}]{joshi-etal-2020-spanbert}
Mandar Joshi, Danqi Chen, Yinhan Liu, Daniel~S. Weld, Luke Zettlemoyer, and Omer Levy. 2020.
\newblock \href {https://doi.org/10.1162/tacl_a_00300} {{S}pan{BERT}: Improving pre-training by representing and predicting spans}.
\newblock \emph{Transactions of the Association for Computational Linguistics}, 8:64--77.

\bibitem[{Kang et~al.(2024)Kang, Lee, Baek, Kawaguchi, and Hwang}]{10.5555/3666122.3668231}
Minki Kang, Seanie Lee, Jinheon Baek, Kenji Kawaguchi, and Sung~Ju Hwang. 2024.
\newblock Knowledge-augmented reasoning distillation for small language models in knowledge-intensive tasks.
\newblock In \emph{Proceedings of the 37th International Conference on Neural Information Processing Systems}, NIPS '23, Red Hook, NY, USA. Curran Associates Inc.

\bibitem[{Karpukhin et~al.(2020)Karpukhin, Oguz, Min, Lewis, Wu, Edunov, Chen, and Yih}]{karpukhin-etal-2020-dense}
Vladimir Karpukhin, Barlas Oguz, Sewon Min, Patrick Lewis, Ledell Wu, Sergey Edunov, Danqi Chen, and Wen-tau Yih. 2020.
\newblock \href {https://doi.org/10.18653/v1/2020.emnlp-main.550} {Dense passage retrieval for open-domain question answering}.
\newblock In \emph{Proceedings of the 2020 Conference on Empirical Methods in Natural Language Processing (EMNLP)}, pages 6769--6781, Online. Association for Computational Linguistics.

\bibitem[{Kim et~al.(2024)Kim, Nam, Mo, Park, Lee, Seo, Ha, and Shin}]{kim2024sure}
Jaehyung Kim, Jaehyun Nam, Sangwoo Mo, Jongjin Park, Sang-Woo Lee, Minjoon Seo, Jung-Woo Ha, and Jinwoo Shin. 2024.
\newblock \href {https://openreview.net/forum?id=w4DW6qkRmt} {Sure: Summarizing retrievals using answer candidates for open-domain {QA} of {LLM}s}.
\newblock In \emph{The Twelfth International Conference on Learning Representations}.

\bibitem[{Krithara et~al.(2023)Krithara, Nentidis, Bougiatiotis, and Paliouras}]{krithara2023bioasq}
Anastasia Krithara, Anastasios Nentidis, Konstantinos Bougiatiotis, and Georgios Paliouras. 2023.
\newblock Bioasq-qa: A manually curated corpus for biomedical question answering.
\newblock \emph{Scientific Data}, 10(1):170.

\bibitem[{Kwon et~al.(2023)Kwon, Li, Zhuang, Sheng, Zheng, Yu, Gonzalez, Zhang, and Stoica}]{10.1145/3600006.3613165}
Woosuk Kwon, Zhuohan Li, Siyuan Zhuang, Ying Sheng, Lianmin Zheng, Cody~Hao Yu, Joseph Gonzalez, Hao Zhang, and Ion Stoica. 2023.
\newblock \href {https://doi.org/10.1145/3600006.3613165} {Efficient memory management for large language model serving with pagedattention}.
\newblock In \emph{Proceedings of the 29th Symposium on Operating Systems Principles}, SOSP '23, page 611–626, New York, NY, USA. Association for Computing Machinery.

\bibitem[{Lassance and Clinchant(2022)}]{10.1145/3477495.3531833}
Carlos Lassance and St\'{e}phane Clinchant. 2022.
\newblock \href {https://doi.org/10.1145/3477495.3531833} {An efficiency study for splade models}.
\newblock In \emph{Proceedings of the 45th International ACM SIGIR Conference on Research and Development in Information Retrieval}, SIGIR '22, page 2220–2226, New York, NY, USA. Association for Computing Machinery.

\bibitem[{Lewis et~al.(2020)Lewis, Liu, Goyal, Ghazvininejad, Mohamed, Levy, Stoyanov, and Zettlemoyer}]{lewis-etal-2020-bart}
Mike Lewis, Yinhan Liu, Naman Goyal, Marjan Ghazvininejad, Abdelrahman Mohamed, Omer Levy, Veselin Stoyanov, and Luke Zettlemoyer. 2020.
\newblock \href {https://doi.org/10.18653/v1/2020.acl-main.703} {{BART}: Denoising sequence-to-sequence pre-training for natural language generation, translation, and comprehension}.
\newblock In \emph{Proceedings of the 58th Annual Meeting of the Association for Computational Linguistics}, pages 7871--7880, Online. Association for Computational Linguistics.

\bibitem[{Li et~al.(2023)Li, Dong, Guerin, and Lin}]{li-etal-2023-compressing}
Yucheng Li, Bo~Dong, Frank Guerin, and Chenghua Lin. 2023.
\newblock \href {https://doi.org/10.18653/v1/2023.emnlp-main.391} {Compressing context to enhance inference efficiency of large language models}.
\newblock In \emph{Proceedings of the 2023 Conference on Empirical Methods in Natural Language Processing}, pages 6342--6353, Singapore. Association for Computational Linguistics.

\bibitem[{Lin(2004)}]{lin-2004-rouge}
Chin-Yew Lin. 2004.
\newblock \href {https://aclanthology.org/W04-1013} {{ROUGE}: A package for automatic evaluation of summaries}.
\newblock In \emph{Text Summarization Branches Out}, pages 74--81, Barcelona, Spain. Association for Computational Linguistics.

\bibitem[{Lin et~al.(2024{\natexlab{a}})Lin, Tang, Tang, Yang, Chen, Wang, Xiao, Dang, Gan, and Han}]{MLSYS2024_42a452cb}
Ji~Lin, Jiaming Tang, Haotian Tang, Shang Yang, Wei-Ming Chen, Wei-Chen Wang, Guangxuan Xiao, Xingyu Dang, Chuang Gan, and Song Han. 2024{\natexlab{a}}.
\newblock \href {https://proceedings.mlsys.org/paper_files/paper/2024/file/42a452cbafa9dd64e9ba4aa95cc1ef21-Paper-Conference.pdf} {Awq: Activation-aware weight quantization for on-device llm compression and acceleration}.
\newblock In \emph{Proceedings of Machine Learning and Systems}, volume~6, pages 87--100.

\bibitem[{Lin et~al.(2024{\natexlab{b}})Lin, Chen, Chen, Shi, Lomeli, James, Rodriguez, Kahn, Szilvasy, Lewis, Zettlemoyer, and tau Yih}]{lin2024radit}
Xi~Victoria Lin, Xilun Chen, Mingda Chen, Weijia Shi, Maria Lomeli, Richard James, Pedro Rodriguez, Jacob Kahn, Gergely Szilvasy, Mike Lewis, Luke Zettlemoyer, and Wen tau Yih. 2024{\natexlab{b}}.
\newblock \href {https://openreview.net/forum?id=22OTbutug9} {{RA}-{DIT}: Retrieval-augmented dual instruction tuning}.
\newblock In \emph{The Twelfth International Conference on Learning Representations}.

\bibitem[{Liu et~al.(2024)Liu, Lin, Hewitt, Paranjape, Bevilacqua, Petroni, and Liang}]{liu-etal-2024-lost}
Nelson~F. Liu, Kevin Lin, John Hewitt, Ashwin Paranjape, Michele Bevilacqua, Fabio Petroni, and Percy Liang. 2024.
\newblock \href {https://doi.org/10.1162/tacl_a_00638} {Lost in the middle: How language models use long contexts}.
\newblock \emph{Transactions of the Association for Computational Linguistics}, 12:157--173.

\bibitem[{Liu et~al.(2023{\natexlab{a}})Liu, Cho, Freedman, Ma, and May}]{liu-etal-2023-recap}
Shuai Liu, Hyundong Cho, Marjorie Freedman, Xuezhe Ma, and Jonathan May. 2023{\natexlab{a}}.
\newblock \href {https://doi.org/10.18653/v1/2023.acl-long.468} {{RECAP}: Retrieval-enhanced context-aware prefix encoder for personalized dialogue response generation}.
\newblock In \emph{Proceedings of the 61st Annual Meeting of the Association for Computational Linguistics (Volume 1: Long Papers)}, pages 8404--8419, Toronto, Canada. Association for Computational Linguistics.

\bibitem[{Liu et~al.(2023{\natexlab{b}})Liu, Iter, Xu, Wang, Xu, and Zhu}]{liu-etal-2023-g}
Yang Liu, Dan Iter, Yichong Xu, Shuohang Wang, Ruochen Xu, and Chenguang Zhu. 2023{\natexlab{b}}.
\newblock \href {https://doi.org/10.18653/v1/2023.emnlp-main.153} {{G}-eval: {NLG} evaluation using gpt-4 with better human alignment}.
\newblock In \emph{Proceedings of the 2023 Conference on Empirical Methods in Natural Language Processing}, pages 2511--2522, Singapore. Association for Computational Linguistics.

\bibitem[{Long et~al.(2023)Long, Wang, and Pan}]{long-etal-2023-adapt}
Quanyu Long, Wenya Wang, and Sinno Pan. 2023.
\newblock \href {https://doi.org/10.18653/v1/2023.emnlp-main.402} {Adapt in contexts: Retrieval-augmented domain adaptation via in-context learning}.
\newblock In \emph{Proceedings of the 2023 Conference on Empirical Methods in Natural Language Processing}, pages 6525--6542, Singapore. Association for Computational Linguistics.

\bibitem[{Loshchilov and Hutter(2019)}]{loshchilov2018decoupled}
Ilya Loshchilov and Frank Hutter. 2019.
\newblock \href {https://openreview.net/forum?id=Bkg6RiCqY7} {Decoupled weight decay regularization}.
\newblock In \emph{International Conference on Learning Representations}.

\bibitem[{Louis et~al.(2024)Louis, van Dijck, and Spanakis}]{Louis_van_Dijck_Spanakis_2024}
Antoine Louis, Gijs van Dijck, and Gerasimos Spanakis. 2024.
\newblock \href {https://doi.org/10.1609/aaai.v38i20.30232} {Interpretable long-form legal question answering with retrieval-augmented large language models}.
\newblock \emph{Proceedings of the AAAI Conference on Artificial Intelligence}, 38(20):22266--22275.

\bibitem[{Nan et~al.(2021)Nan, Nallapati, Wang, Nogueira~dos Santos, Zhu, Zhang, McKeown, and Xiang}]{nan-etal-2021-entity}
Feng Nan, Ramesh Nallapati, Zhiguo Wang, Cicero Nogueira~dos Santos, Henghui Zhu, Dejiao Zhang, Kathleen McKeown, and Bing Xiang. 2021.
\newblock \href {https://doi.org/10.18653/v1/2021.eacl-main.235} {Entity-level factual consistency of abstractive text summarization}.
\newblock In \emph{Proceedings of the 16th Conference of the European Chapter of the Association for Computational Linguistics: Main Volume}, pages 2727--2733, Online. Association for Computational Linguistics.

\bibitem[{Neumann et~al.(2019)Neumann, King, Beltagy, and Ammar}]{neumann-etal-2019-scispacy}
Mark Neumann, Daniel King, Iz~Beltagy, and Waleed Ammar. 2019.
\newblock \href {https://doi.org/10.18653/v1/W19-5034} {{S}cispa{C}y: Fast and robust models for biomedical natural language processing}.
\newblock In \emph{Proceedings of the 18th BioNLP Workshop and Shared Task}, pages 319--327, Florence, Italy. Association for Computational Linguistics.

\bibitem[{Nori et~al.(2023)Nori, King, McKinney, Carignan, and Horvitz}]{nori2023capabilitiesgpt4medicalchallenge}
Harsha Nori, Nicholas King, Scott~Mayer McKinney, Dean Carignan, and Eric Horvitz. 2023.
\newblock \href {https://arxiv.org/abs/2303.13375} {Capabilities of gpt-4 on medical challenge problems}.
\newblock \emph{Preprint}, arXiv:2303.13375.

\bibitem[{Nussbaum et~al.(2024)Nussbaum, Morris, Duderstadt, and Mulyar}]{nussbaum2024nomicembedtrainingreproducible}
Zach Nussbaum, John~X. Morris, Brandon Duderstadt, and Andriy Mulyar. 2024.
\newblock \href {https://arxiv.org/abs/2402.01613} {Nomic embed: Training a reproducible long context text embedder}.
\newblock \emph{Preprint}, arXiv:2402.01613.

\bibitem[{OpenAI(2024)}]{gpt4o-mini}
OpenAI. 2024.
\newblock \href {https://openai.com/index/gpt-4o-mini-advancing-cost-efficient-intelligence/} {Gpt-4o mini: advancing cost-efficient intelligence}.

\bibitem[{Raffel et~al.(2020)Raffel, Shazeer, Roberts, Lee, Narang, Matena, Zhou, Li, and Liu}]{JMLR:v21:20-074}
Colin Raffel, Noam Shazeer, Adam Roberts, Katherine Lee, Sharan Narang, Michael Matena, Yanqi Zhou, Wei Li, and Peter~J. Liu. 2020.
\newblock \href {http://jmlr.org/papers/v21/20-074.html} {Exploring the limits of transfer learning with a unified text-to-text transformer}.
\newblock \emph{Journal of Machine Learning Research}, 21(140):1--67.

\bibitem[{Ram et~al.(2023)Ram, Levine, Dalmedigos, Muhlgay, Shashua, Leyton-Brown, and Shoham}]{ram-etal-2023-context}
Ori Ram, Yoav Levine, Itay Dalmedigos, Dor Muhlgay, Amnon Shashua, Kevin Leyton-Brown, and Yoav Shoham. 2023.
\newblock \href {https://doi.org/10.1162/tacl_a_00605} {In-context retrieval-augmented language models}.
\newblock \emph{Transactions of the Association for Computational Linguistics}, 11:1316--1331.

\bibitem[{Ren et~al.(2024)Ren, Zhan, Wu, and Li}]{ren2024empoweringcharacterleveltextinfilling}
Houxing Ren, Mingjie Zhan, Zhongyuan Wu, and Hongsheng Li. 2024.
\newblock \href {https://arxiv.org/abs/2405.17103} {Empowering character-level text infilling by eliminating sub-tokens}.
\newblock \emph{Preprint}, arXiv:2405.17103.

\bibitem[{Robertson et~al.(2009)Robertson, Zaragoza et~al.}]{robertson2009probabilistic}
Stephen Robertson, Hugo Zaragoza, et~al. 2009.
\newblock The probabilistic relevance framework: Bm25 and beyond.
\newblock \emph{Foundations and Trends{\textregistered} in Information Retrieval}, 3(4):333--389.

\bibitem[{Ryu et~al.(2023)Ryu, Lee, Pang, Choi, Choi, Min, and Sohn}]{ryu-etal-2023-retrieval}
Cheol Ryu, Seolhwa Lee, Subeen Pang, Chanyeol Choi, Hojun Choi, Myeonggee Min, and Jy-Yong Sohn. 2023.
\newblock \href {https://doi.org/10.18653/v1/2023.nllp-1.13} {Retrieval-based evaluation for {LLM}s: A case study in {K}orean legal {QA}}.
\newblock In \emph{Proceedings of the Natural Legal Language Processing Workshop 2023}, pages 132--137, Singapore. Association for Computational Linguistics.

\bibitem[{Ryu et~al.(2024)Ryu, Do, Kim, Lee, and Ok}]{ryu2024keyelementinformedsllmtuningdocument}
Sangwon Ryu, Heejin Do, Yunsu Kim, Gary~Geunbae Lee, and Jungseul Ok. 2024.
\newblock \href {https://arxiv.org/abs/2406.04625} {Key-element-informed sllm tuning for document summarization}.
\newblock \emph{Preprint}, arXiv:2406.04625.

\bibitem[{Sarthi et~al.(2024)Sarthi, Abdullah, Tuli, Khanna, Goldie, and Manning}]{sarthi2024raptor}
Parth Sarthi, Salman Abdullah, Aditi Tuli, Shubh Khanna, Anna Goldie, and Christopher~D Manning. 2024.
\newblock \href {https://openreview.net/forum?id=GN921JHCRw} {{RAPTOR}: Recursive abstractive processing for tree-organized retrieval}.
\newblock In \emph{The Twelfth International Conference on Learning Representations}.

\bibitem[{Shi et~al.(2024)Shi, Min, Yasunaga, Seo, James, Lewis, Zettlemoyer, and Yih}]{shi-etal-2024-replug}
Weijia Shi, Sewon Min, Michihiro Yasunaga, Minjoon Seo, Richard James, Mike Lewis, Luke Zettlemoyer, and Wen-tau Yih. 2024.
\newblock \href {https://doi.org/10.18653/v1/2024.naacl-long.463} {{REPLUG}: Retrieval-augmented black-box language models}.
\newblock In \emph{Proceedings of the 2024 Conference of the North American Chapter of the Association for Computational Linguistics: Human Language Technologies (Volume 1: Long Papers)}, pages 8371--8384, Mexico City, Mexico. Association for Computational Linguistics.

\bibitem[{Team et~al.(2024)Team, Mesnard, Hardin, Dadashi, Bhupatiraju, Pathak, Sifre, Rivi{\`e}re, Kale, Love et~al.}]{team2024gemma}
Gemma Team, Thomas Mesnard, Cassidy Hardin, Robert Dadashi, Surya Bhupatiraju, Shreya Pathak, Laurent Sifre, Morgane Rivi{\`e}re, Mihir~Sanjay Kale, Juliette Love, et~al. 2024.
\newblock Gemma: Open models based on gemini research and technology.
\newblock \emph{arXiv preprint arXiv:2403.08295}.

\bibitem[{Touvron et~al.(2023)Touvron, Martin, Stone, Albert, Almahairi, Babaei, Bashlykov, Batra, Bhargava, Bhosale et~al.}]{touvron2023llama}
Hugo Touvron, Louis Martin, Kevin Stone, Peter Albert, Amjad Almahairi, Yasmine Babaei, Nikolay Bashlykov, Soumya Batra, Prajjwal Bhargava, Shruti Bhosale, et~al. 2023.
\newblock Llama 2: Open foundation and fine-tuned chat models.
\newblock \emph{arXiv preprint arXiv:2307.09288}.

\bibitem[{Wang et~al.(2024{\natexlab{a}})Wang, Ivison, Dasigi, Hessel, Khot, Chandu, Wadden, MacMillan, Smith, Beltagy, and Hajishirzi}]{10.5555/3666122.3669390}
Yizhong Wang, Hamish Ivison, Pradeep Dasigi, Jack Hessel, Tushar Khot, Khyathi~Raghavi Chandu, David Wadden, Kelsey MacMillan, Noah~A. Smith, Iz~Beltagy, and Hannaneh Hajishirzi. 2024{\natexlab{a}}.
\newblock How far can camels go? exploring the state of instruction tuning on open resources.
\newblock In \emph{Proceedings of the 37th International Conference on Neural Information Processing Systems}, NIPS '23, Red Hook, NY, USA. Curran Associates Inc.

\bibitem[{Wang et~al.(2024{\natexlab{b}})Wang, Ma, and Chen}]{wang2024augmentingblackboxllmsmedical}
Yubo Wang, Xueguang Ma, and Wenhu Chen. 2024{\natexlab{b}}.
\newblock \href {https://arxiv.org/abs/2309.02233} {Augmenting black-box llms with medical textbooks for clinical question answering}.
\newblock \emph{Preprint}, arXiv:2309.02233.

\bibitem[{Wang et~al.(2024{\natexlab{c}})Wang, Liu, Lin, Li, Ma, and Liang}]{wang2024ratretrievalaugmentedthoughts}
Zihao Wang, Anji Liu, Haowei Lin, Jiaqi Li, Xiaojian Ma, and Yitao Liang. 2024{\natexlab{c}}.
\newblock \href {https://arxiv.org/abs/2403.05313} {Rat: Retrieval augmented thoughts elicit context-aware reasoning in long-horizon generation}.
\newblock \emph{Preprint}, arXiv:2403.05313.

\bibitem[{Wolf et~al.(2020)Wolf, Debut, Sanh, Chaumond, Delangue, Moi, Cistac, Rault, Louf, Funtowicz, Davison, Shleifer, von Platen, Ma, Jernite, Plu, Xu, Scao, Gugger, Drame, Lhoest, and Rush}]{wolf-etal-2020-transformers}
Thomas Wolf, Lysandre Debut, Victor Sanh, Julien Chaumond, Clement Delangue, Anthony Moi, Pierric Cistac, Tim Rault, Rémi Louf, Morgan Funtowicz, Joe Davison, Sam Shleifer, Patrick von Platen, Clara Ma, Yacine Jernite, Julien Plu, Canwen Xu, Teven~Le Scao, Sylvain Gugger, Mariama Drame, Quentin Lhoest, and Alexander~M. Rush. 2020.
\newblock \href {https://www.aclweb.org/anthology/2020.emnlp-demos.6} {Transformers: State-of-the-art natural language processing}.
\newblock In \emph{Proceedings of the 2020 Conference on Empirical Methods in Natural Language Processing: System Demonstrations}, pages 38--45, Online. Association for Computational Linguistics.

\bibitem[{Xiao et~al.(2024)Xiao, Liu, Zhang, Muennighoff, Lian, and Nie}]{10.1145/3626772.3657878}
Shitao Xiao, Zheng Liu, Peitian Zhang, Niklas Muennighoff, Defu Lian, and Jian-Yun Nie. 2024.
\newblock \href {https://doi.org/10.1145/3626772.3657878} {C-pack: Packed resources for general chinese embeddings}.
\newblock In \emph{Proceedings of the 47th International ACM SIGIR Conference on Research and Development in Information Retrieval}, SIGIR '24, page 641–649, New York, NY, USA. Association for Computing Machinery.

\bibitem[{Xiong et~al.(2024)Xiong, Jin, Lu, and Zhang}]{xiong2024benchmarking}
Guangzhi Xiong, Qiao Jin, Zhiyong Lu, and Aidong Zhang. 2024.
\newblock Benchmarking retrieval-augmented generation for medicine.
\newblock \emph{arXiv preprint arXiv:2402.13178}.

\bibitem[{Xu et~al.(2024{\natexlab{a}})Xu, Shi, and Choi}]{xu2024recomp}
Fangyuan Xu, Weijia Shi, and Eunsol Choi. 2024{\natexlab{a}}.
\newblock \href {https://openreview.net/forum?id=mlJLVigNHp} {{RECOMP}: Improving retrieval-augmented {LM}s with context compression and selective augmentation}.
\newblock In \emph{The Twelfth International Conference on Learning Representations}.

\bibitem[{Xu et~al.(2024{\natexlab{b}})Xu, Ping, Wu, McAfee, Zhu, Liu, Subramanian, Bakhturina, Shoeybi, and Catanzaro}]{xu2024retrievalmeetslongcontext}
Peng Xu, Wei Ping, Xianchao Wu, Lawrence McAfee, Chen Zhu, Zihan Liu, Sandeep Subramanian, Evelina Bakhturina, Mohammad Shoeybi, and Bryan Catanzaro. 2024{\natexlab{b}}.
\newblock \href {https://arxiv.org/abs/2310.03025} {Retrieval meets long context large language models}.
\newblock \emph{Preprint}, arXiv:2310.03025.

\bibitem[{Xu et~al.(2023)Xu, Namazifar, Hazarika, Padmakumar, Liu, and Hakkani-Tür}]{https://doi.org/10.48550/arxiv.2302.09170}
Yan Xu, Mahdi Namazifar, Devamanyu Hazarika, Aishwarya Padmakumar, Yang Liu, and Dilek Hakkani-Tür. 2023.
\newblock \href {https://doi.org/10.48550/ARXIV.2302.09170} {Kilm: Knowledge injection into encoder-decoder language models}.
\newblock \emph{arXiv preprint}.

\bibitem[{Yagnik et~al.(2024)Yagnik, Jhaveri, Sharma, and Pila}]{yagnik2024medlmexploringlanguagemodels}
Niraj Yagnik, Jay Jhaveri, Vivek Sharma, and Gabriel Pila. 2024.
\newblock \href {https://arxiv.org/abs/2401.11389} {Medlm: Exploring language models for medical question answering systems}.
\newblock \emph{Preprint}, arXiv:2401.11389.

\bibitem[{Yang et~al.(2023)Yang, Li, Zhang, Wang, Cheng, Li, and Xiao}]{yang-etal-2023-prca}
Haoyan Yang, Zhitao Li, Yong Zhang, Jianzong Wang, Ning Cheng, Ming Li, and Jing Xiao. 2023.
\newblock \href {https://doi.org/10.18653/v1/2023.emnlp-main.326} {{PRCA}: Fitting black-box large language models for retrieval question answering via pluggable reward-driven contextual adapter}.
\newblock In \emph{Proceedings of the 2023 Conference on Empirical Methods in Natural Language Processing}, pages 5364--5375, Singapore. Association for Computational Linguistics.

\bibitem[{Yu et~al.(2023{\natexlab{a}})Yu, Iter, Wang, Xu, Ju, Sanyal, Zhu, Zeng, and Jiang}]{yu2023generate}
Wenhao Yu, Dan Iter, Shuohang Wang, Yichong Xu, Mingxuan Ju, Soumya Sanyal, Chenguang Zhu, Michael Zeng, and Meng Jiang. 2023{\natexlab{a}}.
\newblock \href {https://openreview.net/forum?id=fB0hRu9GZUS} {Generate rather than retrieve: Large language models are strong context generators}.
\newblock In \emph{The Eleventh International Conference on Learning Representations}.

\bibitem[{Yu et~al.(2023{\natexlab{b}})Yu, Zhang, Pan, Ma, Wang, and Yu}]{yu2023chainofnoteenhancingrobustnessretrievalaugmented}
Wenhao Yu, Hongming Zhang, Xiaoman Pan, Kaixin Ma, Hongwei Wang, and Dong Yu. 2023{\natexlab{b}}.
\newblock \href {https://arxiv.org/abs/2311.09210} {Chain-of-note: Enhancing robustness in retrieval-augmented language models}.
\newblock \emph{Preprint}, arXiv:2311.09210.

\bibitem[{Yu et~al.(2023{\natexlab{c}})Yu, Xiong, Yu, and Liu}]{yu-etal-2023-augmentation}
Zichun Yu, Chenyan Xiong, Shi Yu, and Zhiyuan Liu. 2023{\natexlab{c}}.
\newblock \href {https://doi.org/10.18653/v1/2023.acl-long.136} {Augmentation-adapted retriever improves generalization of language models as generic plug-in}.
\newblock In \emph{Proceedings of the 61st Annual Meeting of the Association for Computational Linguistics (Volume 1: Long Papers)}, pages 2421--2436, Toronto, Canada. Association for Computational Linguistics.

\bibitem[{Zhang* et~al.(2020)Zhang*, Kishore*, Wu*, Weinberger, and Artzi}]{bert-score}
Tianyi Zhang*, Varsha Kishore*, Felix Wu*, Kilian~Q. Weinberger, and Yoav Artzi. 2020.
\newblock \href {https://openreview.net/forum?id=SkeHuCVFDr} {Bertscore: Evaluating text generation with bert}.
\newblock In \emph{International Conference on Learning Representations}.

\bibitem[{Zhong et~al.(2022)Zhong, Liu, Yin, Mao, Jiao, Liu, Zhu, Ji, and Han}]{zhong-etal-2022-towards}
Ming Zhong, Yang Liu, Da~Yin, Yuning Mao, Yizhu Jiao, Pengfei Liu, Chenguang Zhu, Heng Ji, and Jiawei Han. 2022.
\newblock \href {https://doi.org/10.18653/v1/2022.emnlp-main.131} {Towards a unified multi-dimensional evaluator for text generation}.
\newblock In \emph{Proceedings of the 2022 Conference on Empirical Methods in Natural Language Processing}, pages 2023--2038, Abu Dhabi, United Arab Emirates. Association for Computational Linguistics.

\bibitem[{Zhu et~al.(2020)Zhu, Ahuja, Juan, Wei, and Reddy}]{zhu-etal-2020-question}
Ming Zhu, Aman Ahuja, Da-Cheng Juan, Wei Wei, and Chandan~K. Reddy. 2020.
\newblock \href {https://doi.org/10.18653/v1/2020.findings-emnlp.342} {Question answering with long multiple-span answers}.
\newblock In \emph{Findings of the Association for Computational Linguistics: EMNLP 2020}, pages 3840--3849, Online. Association for Computational Linguistics.

\end{thebibliography}
\clearpage
\appendix

\section{Experimental Details}
\label{sec:exp_detail}
\subsection{Model Details}
\label{sec:model_detail}
To demonstrate that our contribution works universally regardless of reader models, we used a range of large language models (LLMs)~\citep{llama3modelcard, jiang2024mixtral, han2023medalpaca, chen2023meditron}, including a commercial model~\citep{hurst2024gpt}, with varying parameters and purposes. All open-source models were implemented based on HuggingFace's Transformers~\citep{wolf-etal-2020-transformers}, and due to hardware constraints, we utilized AWQ~\citep{MLSYS2024_42a452cb} models for LLMs with a substantial number of parameters. The specific model details are as follows:
\begin{table}[h]
\resizebox{\columnwidth}{!}{%
\begin{tabular}{l|l}
\toprule
\textbf{Model} & \textbf{HuggingFace's Repository} \\
\midrule
\textsc{K-comp} & google/gemma-2b \\
Retriever & nomic-ai/nomic-embed-text-v1.5 \\
Llama-3 8B & meta-llama/Meta-Llama-3-8B \\
Llama-3 70B & TechxGenus/Meta-Llama-3-70B-AWQ \\
Mixtral-8x7B & TheBloke/mixtral-8x7b-v0.1-AWQ \\
MedAlpaca 13B & medalpaca/medalpaca-13b \\
Meditron 70B & TheBloke/meditron-70B-AWQ \\
Selective-Context & meta-llama/Llama-2-7b-hf \\
LLMLingua & meta-llama/Llama-2-7b-hf \\
SPLADE & naver/efficient-splade-V-large-query \\
 & naver/efficient-splade-V-large-doc \\
\bottomrule
\end{tabular}%
}
\end{table}

\subsection{Dataset Details}
The datasets employed in the main experiments are MedQuAD~\citep{BenAbacha-BMC-2019}, MASH-QA~\citep{zhu-etal-2020-question}, and BioASQ~\citep{krithara2023bioasq}. MedQuAD encompasses a wide range of question types related to biomedicine, such as diseases, drugs, and medical tests. MASH-QA is a dataset from the consumer health domain where answers need to be extracted from multiple, non-consecutive parts of a long document. BioASQ is a biomedical dataset derived from PubMed, designed to support a range of tasks, including question-answering, information retrieval, and summarization. We obtained each dataset from the official websites provided by the papers (e.g., GitHub). In the case of the MedQuAD dataset, since there is no test data available, we randomly split the dataset into train/validation/test sets with an 80/10/10 ratio to conduct our experiments. Additionally, we used the MEDIQA~\citep{ben-abacha-etal-2019-overview} for further comparison with the baseline in GPT-4o evaluation section. MEDIQA comprises three tasks: Natural Language Inference, Recognizing Question Entailment, and QA, but only the QA task was used.

\subsection{Implementation Details}
\label{sec:imple_datail}
We fine-tuned Gemma-2B~\citep{team2024gemma} with our knowledge injection objective and also used the FineTune baseline (FineTune) with the same model. Table~\ref{tab:hyperparam} shows the hyperparameter settings that were used for training and inference. \textsc{K-comp} and FineTune were trained with the same hyperparameters and selected their optimal checkpoints based on the performance of the development set. We also trained the model via the Transforming Reinforcement Learning~\citep{hu2024transforming}. When training on MASH-QA, we used 4 NVIDIA A100 GPUs with 80GB memory, otherwise we used 2 A100-80GB GPUs. For inference, we conducted experiments on a single A100-80GB GPU with vLLM~\citep{10.1145/3600006.3613165} to accelerate and efficiently perform inference.

\begin{table*}[t]
\resizebox{\textwidth}{!}{%
\begin{tabular}{lcccccccccccc}
\toprule
 & \multicolumn{2}{c}{\textbf{Llama-3-8B}} & \multicolumn{2}{c}{\textbf{Llama-3-70B}} & \multicolumn{2}{c}{\textbf{Mixtral-8x7B}} & \multicolumn{2}{c}{\textbf{GPT-4o}} & \multicolumn{2}{c}{\textbf{MedAlpaca-13B}} & \multicolumn{2}{c}{\textbf{Meditron-70B}} \\
 & \textbf{BertScore} & \textbf{UniEval} & \textbf{BertScore} & \textbf{UniEval} & \textbf{BertScore} & \textbf{UniEval} & \textbf{BertScore} & \textbf{UniEval} & \textbf{BertScore} & \textbf{UniEval} & \textbf{BertScore} & \textbf{UniEval} \\ \midrule
\multicolumn{13}{c}{\underline{MedQuAD}} \\
RAPTOR & 74.96 & 60.84 & 77.66 & \textbf{71.79} & 66.4 & 61.49 & 84.20 & 57.78 & 83.59 & 48.42 & 74.67 & \textbf{60.30} \\
\rowcolor[HTML]{EFEFEF}
\textsc{K-comp} & \textbf{83.79} & \textbf{64.05} & \textbf{82.06} & 62.91 & \textbf{80.64} & \textbf{63.07} & \textbf{85.49} & \textbf{63.16} & \textbf{85.75} & \textbf{49.39} & \textbf{78.88} & 58.95 \\ \midrule
\multicolumn{13}{c}{\underline{MASH-QA}} \\
RAPTOR & 79.61 & 60.62 & 84.64 & \textbf{66.20} & 73.37 & \textbf{62.95} & 83.31 & 58.60 & 80.19 & \textbf{50.27} & 80.49 & \textbf{64.83} \\
\rowcolor[HTML]{EFEFEF}
\textsc{K-comp} & \textbf{83.45} & \textbf{61.45} & \textbf{83.64} & 61.40 & \textbf{81.49} & 62.91 & \textbf{84.49} & \textbf{64.06} & \textbf{81.95} & 51.76 & \textbf{82.67} & 61.85 \\ \midrule
\multicolumn{13}{c}{\underline{BioASQ}} \\
RAPTOR & 82.63 & 60.62 & 85.57 & \textbf{70.32} & 83.39 & 66.10 & 87.16 & 56.64 & 83.50 & \textbf{46.40} & 85.44 & \textbf{65.48} \\
\rowcolor[HTML]{EFEFEF}
\textsc{K-comp} & \textbf{86.61} & \textbf{62.28} & \textbf{87.03} & 62.89 & \textbf{86.52} & \textbf{66.31} & \textbf{87.77} & \textbf{64.38} & \textbf{84.88} & 45.47 & \textbf{86.70} & 62.90 \\ \bottomrule 
\end{tabular}%
}
\vspace{-1mm}
\caption{\label{tab:raptor_main}RAPTOR results. The implementation follows the same approach as that utilized for Table~\ref{tab:main_result}.}
\vspace{-1mm}
\end{table*}
\section{Analysis of Query-Agnostic}
\label{sec:query-agnostic}
Figure~\ref{fig:gpt_eval_baseline} compares our approach to previous studies that compress prompts without relying on the query, as illustrated in the top-left panel. GPT-4o-mini is the result of summarizing only passages without questions. We referred to the prompts used when training RECOMP, which can be seen in Table~\ref{tab:summary_generation_mini}. Selective-Context (SC)~\citep{li-etal-2023-compressing} is a method for the removal of contexts with low self-information at the token level. SC was compressed with Llama-2-7B~\citep{touvron2023llama} as that used for LLMLingua. Both methods compress prompts by considering token-level dependencies, so the resulting sentences are incomplete when decoding. As can be seen in Table~\ref{tab:example_baseline}, LLMLingua generates sequences in which words and symbols are merely listed. At the same time, SC produces fragments due to token-level pruning, which also results in incomplete sentences. Consequently, although embedding-based metrics such as BertScore~\citep{bert-score} or lexical-based metrics like ROUGE~\citep{lin-2004-rouge} may exhibit robust performance, they are limited in qualitative evaluations. Moreover, the methodology of compressing passages without referencing the query is unsuitable for retrieval-augmented QA tasks. Thus, both GPT-4o-mini and \textsc{K-comp} outperform the aforementioned baselines. In addition, it is expected that \textsc{K-comp}, by referencing masked questions, provides a more relevant and specialized context compared to GPT-4o-mini, which does not reference the questions.

\begin{table}[t]
\small
\centering
\resizebox{\columnwidth}{!}{%
\begin{tabular}{lccc}
\toprule
 & \multicolumn{1}{l}{\textbf{MedQuAD}} & \multicolumn{1}{l}{\textbf{MASH-QA}} & \multicolumn{1}{l}{\textbf{BioASQ}} \\ \midrule
RAPTOR & 20.7\% & 30.1\% & 43.4\% \\
\textsc{K-comp} & 77.9\% & 69.6\% & 52.5\% \\
Tie & 1.4\% & 0.3\% & 4.1\% \\ \bottomrule
\end{tabular}%
}
\vspace{-1mm}
\caption{\label{tab:raptor_gpt_eval}GPT-4o evaluation results with the RAPTOR approach.}
\vspace{-1mm}
\end{table}
\section{Additional Baseline}
In Section~\ref{sec:experiments}, we compared various baselines for RAG methods that incorporate prompt compression. However, prior RAG approaches that do not utilize compression were not implemented. Although the primary focus of our study is not to exhaustively explore non-compression-based methods, we include additional state-of-the-art baselines to facilitate a more comprehensive evaluation of how \textsc{K-comp} performs relative to other approaches. Specifically, we adopt RAPTOR~\citep{sarthi2024raptor} as a baseline and conduct experiments under several predefined conditions. Firstly, RAPTOR requires a summarizing model because it recursively summarizes to construct a tree structure. Therefore, we use FineTune model to perform context summarization, thereby eliminating the penalty for training. Second, due to computational resource limitations, we imposed a size constraint on the retrieval corpus to enable summarization. Given the massive scale of the MedCorp~\citep{xiong2024benchmarking} corpus, we replace it with a smaller corpus consisting of the top 5 documents retrieved by the retriever across the entire dataset. We argue that these experimental conditions do not put RAPTOR at a disadvantage. Rather, we believe that restricting the corpus to the top-5 retrieved documents may offer potential advantages by improving the relevance and quality of the retrieved information.

As shown in Table~\ref{tab:raptor_main}, \textsc{K-comp} proves effective in QA tasks, whether using prompt compression or non-prompt compression approaches. Furthermore, as evidenced in Table~\ref{tab:inference_speed}, \textsc{K-comp} further underscores its cost efficiency, particularly compared to non-prompt compression methods. We also performed the GPT-4o evaluation described in Section~\ref{sec:experiments}, with the results presented in Table~\ref{tab:raptor_gpt_eval}. The results indicate that the passages retrieved by RAPTOR approach are preferred more frequently in BioASQ than other datasets. As seen in Table~\ref{tab:main_result}, the FineTune model is particularly well-tuned for BioASQ, suggesting that RAPTOR effectively integrates both high-level concepts and detailed information during its tree construction process.

Despite RAPTOR's advantage of using a corpus limited to the top-5 documents, \textsc{K-comp} achieves higher accuracy than the state-of-the-art retrieval-augmented QA baseline (non-prompt compression). This validates our hypothesis that knowledge injection enhances QA performance. However, it is worth noting that RAPTOR was originally designed using closed models from the GPT family, which may not align as effectively with our work, particularly when using smaller models such as Gemma.

\section{Licenses}
MedQuAD, MASH-QA, and BioASQ are licensed under CC-BY-4.0, Apache-2.0, and CC-BY-2.5, respectively. The models Gemma-2b, Nomic-embed-text-v1.5, Meta-Llama-3, Mixtral-8x7b-v0.1-AWQ, GPT-4o, MedAlpaca-13b, Meditron-70B-AWQ, and Bge-reranker-large are licensed under Gemma, Apache-2.0, Llama 3 Community, Apache-2.0, OpenAI, Creative Commons, Llama 2 Community, and MIT, respectively.

\onecolumn
\section{Examples of Used Prompts}
\label{sec:appendix}

\begin{table*}[!ht]
\centering
\renewcommand{\arraystretch}{1.25}
\setlength{\tabcolsep}{2pt}
\footnotesize
\begin{tabular}{p{\textwidth}}
\toprule
Please extract the content about the entity in fewer than four sentences.\\\\
\#\#\# Passage \\
Therapies in Aicardi-Goutières syndrome.\\
Aicardi-Goutières syndrome (AGS) is a genetically determined disorder, affecting most particularly the brain and the skin, characterized by the inappropriate induction of a type I interferon-mediated immune response. In most, but not all, cases the condition is severe, with a high associated morbidity and mortality ...(skip)\\\\
Treatments in Aicardi-Goutières syndrome.\\
Comprehensive reviews of the clinical characteristics and pathogenesis of Aicardi-Goutières syndrome (AGS), particularly its contextualization within a putative type I interferonopathy framework, already exist. However, recent reports of attempts at treatment suggest that an assessment of the field from a therapeutic perspective is warranted at this time ...(skip)\\\\
Novel and emerging treatments for Aicardi-Goutières syndrome.\\
<bIntroduction</b: Aicardi-Goutières syndrome (AGS) is the prototype of the type I interferonopathies, a new heterogeneous group of autoinflammatory disorders in which type I interferon plays a pivotal role. The disease usually manifests itself during infancy, primarily affecting the brain and the skin, and is characterized by cerebrospinal fluid chronic lymphocytosis and raised levels of interferon-alpha and by cardinal neuroradiological features: cerebral calcification, leukoencephalopathy and cerebral atrophy ...(skip)\\\\
Aicardi–Goutières syndrome\\
At the moment there are no therapies specifically targeting the underlying cause of AGS. Current treatments address the symptoms, which can be varied both in scope and severity. Many patients benefit from tube-feeding. Drugs can be administered to help with seizures / epilepsy ...(skip)\\\\
Aicardi–Goutières syndrome\\
Treatment\\\\
\#\#\# Entity\\
\text{[research, clinical trial, Disorder]}\\
\bottomrule
\end{tabular}
\caption{Prompt for synthesizing summaries used in \textsc{K-comp}.}
\label{tab:summary_generation}
\end{table*}

\clearpage

\begin{table*}[]
\centering
\renewcommand{\arraystretch}{1.25}
\setlength{\tabcolsep}{2pt}
\footnotesize
\begin{tabular}{p{\textwidth}}
\toprule
Compress the information in the retrieved documents into a 2-sentence summary that could be used to answer the question:\\
Question: what research (or clinical trials) is being done for Aicardi-Goutieres Syndrome Disorde ?\\
Retrieved documents: Therapies in Aicardi-Goutières syndrome.\\
Aicardi-Goutières syndrome (AGS) is a genetically determined disorder, affecting most particularly the brain and the skin, characterized by the inappropriate induction of a type I interferon-mediated immune response. In most, but not all, cases the condition is severe, with a high associated morbidity and mortality ...(skip)\\\\
Treatments in Aicardi-Goutières syndrome.\\
Comprehensive reviews of the clinical characteristics and pathogenesis of Aicardi-Goutières syndrome (AGS), particularly its contextualization within a putative type I interferonopathy framework, already exist. However, recent reports of attempts at treatment suggest that an assessment of the field from a therapeutic perspective is warranted at this time ...(skip)\\\\
Novel and emerging treatments for Aicardi-Goutières syndrome.\\
<bIntroduction</b: Aicardi-Goutières syndrome (AGS) is the prototype of the type I interferonopathies, a new heterogeneous group of autoinflammatory disorders in which type I interferon plays a pivotal role. The disease usually manifests itself during infancy, primarily affecting the brain and the skin, and is characterized by cerebrospinal fluid chronic lymphocytosis and raised levels of interferon-alpha and by cardinal neuroradiological features: cerebral calcification, leukoencephalopathy and cerebral atrophy ...(skip)\\\\
Aicardi–Goutières syndrome\\
At the moment there are no therapies specifically targeting the underlying cause of AGS. Current treatments address the symptoms, which can be varied both in scope and severity. Many patients benefit from tube-feeding. Drugs can be administered to help with seizures / epilepsy ...(skip)\\\\
Aicardi–Goutières syndrome\\
Treatment\\\\
Compressed documents:\\
\bottomrule
\end{tabular}
\vspace{-1mm}
\caption{\label{tab:summary_generation_recomp}Prompt for synthesizing summaries used in RECOMP training and query-based GPT-4o-mini.}

\mbox{}\\
\mbox{}\\
\mbox{}\\
\mbox{}\\
\mbox{}\\

\centering
\renewcommand{\arraystretch}{1.25}
\setlength{\tabcolsep}{2pt}
\footnotesize
\begin{tabular}{p{\textwidth}}
\toprule
Compress the information in the retrieved documents into a 2-sentence summary\\
Retrieved documents: \texttt{\{\{Top-5 retrieved passages\}\}}\\
Compressed documents:\\
\bottomrule
\end{tabular}
\vspace{-1mm}
\caption{Prompt for synthesizing summaries used in query-agnostic GPT-4o-mini.}
\label{tab:summary_generation_mini}
\end{table*}
\clearpage
\begin{table*}[]
\renewcommand{\arraystretch}{1.25}
\setlength{\tabcolsep}{2pt}
\footnotesize
\centering
\begin{tabular}{p{\textwidth}}
\toprule
\#\#\# Passage\\
Aicardi-Goutières syndrome (AGS) is a genetically determined disorder affecting the brain and skin, characterized by inappropriate immune responses due to type I interferon. Current research focuses on understanding its pathogenesis and developing targeted therapies, with some recent attempts exploring treatments like Janus kinase inhibitors and anti-IFN-$\alpha$ antibodies. Despite advancements, there are still challenges in assessing efficacy and addressing open questions related to treatment effectiveness. Ongoing clinical trials aim to evaluate the efficacy of new therapies and address the underlying causes of AGS. Overall, ongoing research and clinical trials are essential for developing effective treatments for AGS and type I interferonopathies.\\\\
\#\#\# Entity\\
research: Systematic study undertaken to increase knowledge \\
clinical trial: Phase of clinical research in medicine\\
Aicardi-Goutières Syndrome: Aicardi-Goutières syndrome is a disorder that mainly affects the brain, the immune system, and the skin.Most newborns with Aicardi-Goutières syndrome do not show any signs or symptoms of the disorder\\\\
\#\#\# Questions \\
what research (or clinical trials) is being done for Aicardi-Goutieres Syndrome Disorde ?\\
\midrule
\#\#\# Passage\\
Therapies in Aicardi-Goutières syndrome.\\
Aicardi-Goutières syndrome (AGS) is a genetically determined disorder, affecting most particularly the brain and the skin, characterized by the inappropriate induction of a type I interferon-mediated immune response. In most, but not all, cases the condition is severe, with a high associated morbidity and mortality ...(skip)\\\\
Treatments in Aicardi-Goutières syndrome.\\
Comprehensive reviews of the clinical characteristics and pathogenesis of Aicardi-Goutières syndrome (AGS), particularly its contextualization within a putative type I interferonopathy framework, already exist. However, recent reports of attempts at treatment suggest that an assessment of the field from a therapeutic perspective is warranted at this time ...(skip)\\\\
Novel and emerging treatments for Aicardi-Goutières syndrome.\\
<bIntroduction</b: Aicardi-Goutières syndrome (AGS) is the prototype of the type I interferonopathies, a new heterogeneous group of autoinflammatory disorders in which type I interferon plays a pivotal role. The disease usually manifests itself during infancy, primarily affecting the brain and the skin, and is characterized by cerebrospinal fluid chronic lymphocytosis and raised levels of interferon-alpha and by cardinal neuroradiological features: cerebral calcification, leukoencephalopathy and cerebral atrophy ...(skip)\\\\
Aicardi–Goutières syndrome\\
At the moment there are no therapies specifically targeting the underlying cause of AGS. Current treatments address the symptoms, which can be varied both in scope and severity. Many patients benefit from tube-feeding. Drugs can be administered to help with seizures / epilepsy ...(skip)\\\\
Aicardi–Goutières syndrome\\
Treatment\\\\
\#\#\# Questions \\
what research (or clinical trials) is being done for Aicardi-Goutieres Syndrome Disorde ?\\
\bottomrule
 \end{tabular}
 \vspace{-1mm}
    \caption{Prompt for reader LLMs. (Above: \textsc{K-comp}, Below: Top-5 passages)}
    \label{tab:reader_prompt}
\end{table*}
\clearpage
\begin{table*}[]
\renewcommand{\arraystretch}{1.25}
\setlength{\tabcolsep}{2pt}
\footnotesize
\centering
\begin{tabular}{p{\textwidth}}
\toprule
Select the summary (Summary 1, Summary 2, Summary 3, or Summary 4) that is more relevant and informative as a rationale for the given question.\\
In particular, biomedical QA requires expertise to be credible, so choose a summary where expertise exists in the domain.\\
Choice: [Summary 1, Summary 2, Summary 3, Summary 4, Tie], Do not offer any opinions other than the choice.\\\\
\#\#\# Summary 1\\
\texttt{\{\{summary 1\}\}}\\\\
\#\#\# Summary 2\\
\texttt{\{\{summary 2\}\}}\\\\
\#\#\# Summary 3\\
\texttt{\{\{summary 3\}\}}\\\\
\#\#\# Summary 4\\
\texttt{\{\{summary 4\}\}}\\\\
\#\#\# Question\\
\texttt{\{\{question\}\}}
\\
\bottomrule
 \end{tabular}
\caption{Prompt for GPT-4o evaluation. The order of summaries was randomized throughout the evaluation process to mitigate potential bias associated with their positioning.}
\label{tab:gpt_eval}
\end{table*}
\clearpage
\begin{table*}[]
\renewcommand{\arraystretch}{1.25}
\setlength{\tabcolsep}{2pt}
\footnotesize
\centering
\begin{tabular}{p{\textwidth}}
\toprule
You will be given a question, passage, and answer (both the provided answer and the correct answer).\\
Your task is to evaluate how coherent and relevant the provided answer is based on the question, passage, and correct answer. The passage is provided as a reference to determine if the provided answer appropriately aligns with the information required to answer the question.\\
Please make sure you read and understand these instructions carefully. Please keep this document open while reviewing, and refer to it as needed.\\\\

Evaluation Criteria:\\
Coherence (1-5) - the degree to which the provided answer is logically consistent, well-structured, and directly relevant to the question. The answer should make logical sense and align with the correct answer, while addressing the question clearly and effectively.\\\\

Evaluation Steps:\\
1. Read the question carefully to identify the specific information it seeks.\\
2. Read the answer and compare it to the correct answer. Check if the provided answer addresses the question directly, aligns logically with the passage's content, and is presented in a clear, organized, and complete manner.\\
3. Assign a score for coherence on a scale of 1 to 5, where 1 is the lowest and 5 is the highest based on the Evaluation Criteria.\\\\

Example:\\
Passage:\\
\texttt{\{\{passage\}\}}\\\\

Question:\\
\texttt{\{\{question\}\}}\\\\

Correct Answer:\\
\texttt{\{\{correct answer\}\}}\\\\

Answer:\\
\texttt{\{\{answer\}\}}\\\\

Evaluation Form (scores ONLY):\\
Coherence:
\\
\bottomrule
 \end{tabular}
\caption{\label{tab:g_eval}Prompt for G-Eval-4 evaluation metric~\citep{liu-etal-2023-g}.}
\end{table*}
\clearpage
\section{Extended Case Study}
\begin{table*}[!ht]
\renewcommand{\arraystretch}{1.25}
\setlength{\tabcolsep}{2pt}
\footnotesize
\centering
\begin{tabular}{p{\textwidth}}
\toprule
\textbf{\textsc{K-comp}}\\\\
\#\#\# Entity\\
x-ray: Form of short-wavelength electromagnetic radiation\\
rheumatoid arthritis: Type of autoimmune arthritis\\\\
\#\#\# Summary\\
X-rays are the most common method for assessing the degree of joint destruction in rheumatoid arthritis (RA), revealing osteopenia and joint space narrowing as the disease progresses. They are particularly useful for identifying bony erosions and assessing treatment efficacy. While imaging techniques like MRI and ultrasound are also employed, X-rays remain the most common for evaluating joint damage in RA. Early diagnosis and treatment are crucial to modify the disease's course and reduce complications. Understanding the disease's pathophysiology and imaging findings is essential for effective diagnosis and management.\\\\

\textbf{Llama-3-8B's Answer} \\
X-rays diagnose rheumatoid arthritis (RA) by revealing changes in the axial skeleton and joint space. As the disease progresses, early RA may not show significant changes on X-rays, but it can indicate osteopenia and joint space narrowing. Advances in imaging techniques like MRI and ultrasound enhance the ability to detect soft tissue changes and joint damage, which is crucial for accurate diagnosis and treatment management. Overall, X-rays serve as an important tool for evaluating the structural changes associated with RA.\\\\

\textbf{Llama-3-70B's Answer} \\
X-rays can reveal joint space narrowing and osteopenia, which are common findings in RA.\\
X-rays can show early signs of joint damage in RA, such as bone erosion and synovial inflammation.\\
X-rays can detect soft tissue changes and joint damage in RA, which is important for diagnosis and treatment management.\\
X-rays can assess the degree of joint destruction in RA, including changes in the axial skeleton and joint space.
\\\\
\midrule
\textbf{FineTune}\\\\
\#\#\# Summary\\
Rheumatoid arthritis (RA) is a chronic systemic disease characterized by inflammation of the synovial tissue, affecting 0.5\%-1.0\% of the global population, with a higher prevalence in females. Modern diagnostic techniques, including X-rays, MRI, and ultrasound, are crucial for assessing joint damage and evaluating treatment efficacy. Early diagnosis and treatment are essential to modify the disease's course and reduce severe complications. Advances in imaging techniques, particularly MRI, enhance the ability to detect and manage RA more effectively. Understanding the disease's pathophysiology and various imaging findings is essential for accurate diagnosis and tailored treatment strategies. Overall, imaging plays a significant role in the management and monitoring of RA.\\\\

\textbf{Llama-3-8B's Answer} \\
X-rays are used to diagnose RA by detecting joint damage and erosion.\\\\

\textbf{Llama-3-70B's Answer} \\
X-rays are a valuable tool in the diagnosis of RA, as they can detect joint damage and erosions caused by the disease. In the early stages of RA, X-rays may not show any abnormalities, but as the disease progresses, they can reveal joint space narrowing, erosions, and deformities. X-rays can also help monitor the progression of the disease and assess the effectiveness of treatment. However, X-rays have limitations, as they may not detect early changes in the joints or soft tissue inflammation. Therefore, other imaging modalities, such as MRI and ultrasound, may be necessary to provide a more comprehensive assessment of RA.
\\
\bottomrule
\end{tabular}
\vspace{-1mm}
\caption{Case study in section 4. (Above: \textsc{K-comp}, Below: FineTune)}
\label{tab:full_case_study}
\end{table*}
\clearpage
\section{Dataset Statistics}
\begin{table*}[h]
\small
\centering
\begin{tabular}{l|ccc|ccc|ccc}
\toprule
 & \multicolumn{3}{c|}{\textbf{MedQuAD}} & \multicolumn{3}{c|}{\textbf{MASH-QA}} & \multicolumn{3}{c}{\textbf{BioASQ}} \\
 & Train & Validation & Test & Train & Validation & Test & Train & Validation & Test \\ 
\midrule
Original & 13,127 & 1,640 & 1,640 & 27,728 & 3,587 & 3,493 & 3,209 & 803 & 707 \\
After filtering & 9,064 & 1,095 & 1,554 & 20,458 & 2,637 & 3,226 & 2,284 & 563 & 647 \\
\% Filtered & 30.9 & 33.2 & 5.2 & 26.2 & 26.4 & 7.6 & 28.8 & 29.8 & 8.4 \\
\bottomrule
\end{tabular}%
\vspace{-1mm}
\caption{Dataset sizes before and after filtering in the entity recognition step. For test data, filtering is applied exclusively to questions lacking any entities. For other datasets, filtering is additionally conducted for the absence of corresponding descriptions for the recognized entities.}
\label{tab:dataset_statistics}
\end{table*}
\section{Hyperparameters}
\begin{table*}[h]
\small
\centering
\begin{tabular}{l|ccc}
\toprule
\multicolumn{1}{c|}{\textbf{Hyperparameters}} & \textbf{MedQuAD} & \textbf{MASH-QA} & \textbf{BioASQ} \\
\midrule
Global batch size & \multicolumn{3}{c}{8} \\
Gradient accumulate & \multicolumn{3}{c}{1} \\
Learning scheduler & \multicolumn{3}{c}{cosine} \\
Weight decay & \multicolumn{3}{c}{3\%} \\
Epochs & \multicolumn{3}{c}{3} \\
Temperature & \multicolumn{3}{c}{0.01} \\
Top\_p & \multicolumn{3}{c}{1.0} \\
Early stopping & \multicolumn{3}{c}{True (valid loss)} \\
Optimizer & \multicolumn{3}{c}{AdamW \citep{loshchilov2018decoupled}} \\
 & \multicolumn{3}{c}{($\beta_1$=0.9, $\beta_2$=0.999, $\epsilon$=1e-8)} \\
Learning rate & 1e-4 & 1e-4 & 1e-5 \\
\bottomrule
\end{tabular}
\vspace{-1mm}
\caption{Hyperparameters used in the experiments.}
\label{tab:hyperparam}
\end{table*}
\section{GPT-4o Evaluation Results}
\begin{table*}[h]
\small
\centering
\resizebox{\textwidth}{!}{%
\begin{tabular}{llcccccc}
\toprule
\multicolumn{2}{l}{\multirow{2}{*}{\textbf{Method}}} & \multicolumn{3}{c}{\textbf{Seen Data}} & \multicolumn{3}{c}{\textbf{Unseen Data}} \\
\cmidrule(lr){3-5} \cmidrule(lr){6-8}
& & MedQuAD & MASH-QA & BioASQ & \begin{tabular}[c]{@{}c@{}}MEDIQA\\ (t-MedQuAD)\end{tabular} & \begin{tabular}[c]{@{}c@{}}MEDIQA\\ (t-MASH-QA)\end{tabular} & \begin{tabular}[c]{@{}c@{}}MEDIQA\\ (t-BioASQ)\end{tabular} \\
\midrule
\multirow{5}{*}{\textbf{Query-Agnostic}} 
& Selective-Context & 0 & 0 & 3 & 0 & 0 & 0 \\
& LLMLingua & 4 & 10 & 1 & 0 & 0 & 1 \\
& GPT-4o-mini & 259 & 635 & 202 & 25 & 19 & 48 \\
& \textsc{K-comp} & 1290 & 2581 & 441 & 115 & 121 & 91 \\
& Tie & 1 & 0 & 0 & 0 & 0 & 0 \\
\midrule
\multirow{5}{*}{\textbf{Query-Based}} 
& SPLADE & 88 & 371 & 70 & 16 & 17 & 19 \\
& RECOMP & 252 & 406 & 82 & 6 & 8 & 4 \\
& GPT-4o-mini & 504 & 1240 & 253 & 56 & 51 & 71 \\
& \textsc{K-comp} & 710 & 1208 & 242 & 62 & 64 & 46 \\
& Tie & 0 & 1 & 0 & 0 & 0 & 0 \\
\bottomrule
\end{tabular}%
}
\vspace{-1mm}
\caption{All results of the GPT-4o evaluation.}
\label{tab:full_gpt_eval}
\end{table*}
\clearpage
\section{Examples of Baselines}
\begin{table*}[!ht]
\centering
\small
\renewcommand{\arraystretch}{1.0}
\setlength{\tabcolsep}{2pt}
\footnotesize
\begin{tabular}{p{\textwidth}}
\toprule
Question: Are there any adults who shouldn't get the meningococcal vaccine?\\
\midrule
\textbf{Selective-Context}\\
Meningococcal disease Routine primary immunization against who planning Although conjugate vaccines in Since safety to Prevention Invasive disease The disease can rapidly or despite with required household Although cases often Serogroup B accounts but contained Adolescents disease Because in contained they potentially recently containing This T-cell-dependent vaccine that should The Advisory Committee to entering The vaccine also may belong Meningococcal disease through caused While IMD incidence peaks both infants carriage rates often increasing Effective vaccines Because represent cases often have the highest carriage rate efforts in This review describes and characteristics With countries that will also Rates increasing. The burden due available older. Incorporation into offers affect The routine use will Awareness attitudes among among Meningococcal- under However MenB vaccination coverage We investigated MenB disease among An online survey parents among There were 606 total participants parents of MenB (<in</i = also Non-vaccinated cohorts reported ( though However all cohorts reported high interest MenB disease Vaccinated cohorts relied had SCDM at though those Barriers included PCP recommendation uncertainty There gaps MenB disease SCDM patients resulting discussing Targeted education on MenB may these opportunities\\\\
\textbf{LLMLingua}\\
Men
ine most or are. Althoughss conjug. Since established

disease
In be ande.otic is for. population. B the.s Becauseoccaloups are the, are The This and The to who to certain the isidence in, are,., haveination, regardingterm of in ofents. on with have
ofis The limitations been populations In more TheA of and3old under population regardingold of (<ini andination, and SCDM among parents and patients in the US, resulting in missed opportunities for discussing and administering MenB vaccination. Targeted education on MenB and vaccination recommendations may increase these opportunities and improve MenB vaccination awareness and initiation.\\\\
\textbf{GPT-4o-mini (query-agnostic)}\\
Routine primary immunization against meningococcal disease is advised for adults in endemic areas or traveling there, with conjugate vaccines preferred for those under 55, while polysaccharide vaccines are used for older adults due to safety concerns. Adolescents and young adults are prioritized for vaccination due to their higher incidence and carriage rates of the disease, yet awareness and vaccination coverage for serogroup B remains low, necessitating targeted education to improve understanding and uptake.\\\\
\textbf{SPLADE}\\
Meningococcal disease
Routine primary immunization against meningococcal disease is recommended for most adults living in areas where meningococcal disease is endemic or who are planning to travel to such areas. Although conjugate vaccines are the preferred meningococcal vaccine in adults 55 years of age or younger, polysaccharide vaccines are an acceptable alternative for adults in this age group if the conjugated vaccine is unavailable. Since safety and efficacy of conjugate vaccines in adults older than 55 years of age have not been established to date, polysaccharide vaccines should be used for primary immunization in this group.\\\\
\textbf{RECOMP}\\
Adults who should not receive the meningococcal vaccine include those who are 55 years or older, those in high-risk groups, and those who are pregnant or breastfeeding. Additionally, polysaccharide vaccines are recommended for adults over 55, as conjugate vaccines are not recommended for this age group due to safety concerns.\\\\
\textbf{GPT-4o-mini (query-based)}\\
Most adults are recommended to receive the meningococcal vaccine, particularly those living in endemic areas or traveling to such regions, but adults over 55 years should use polysaccharide vaccines due to the unestablished safety and efficacy of conjugate vaccines for this age group. Overall, while there are no strict categories of adults who should not get the vaccine, individuals with specific health concerns or those who have experienced adverse reactions to prior vaccinations should consult with healthcare professionals for personalized guidance.\\\\
\textbf{\textsc{K-comp}}\\
\#\#\# Entity\\
people: Plurality of persons considered as a whole\\
meningococcal vaccine: Vaccine against meningococcal disease\\\\
\#\#\# Summary\\
The meningococcal vaccine is recommended for adults in endemic areas and for adolescents and young adults in high-risk groups. The meningococcal vaccine is safe and effective, but the polysaccharide vaccine is preferred for those 55 years and older due to safety concerns in older adults. Vaccination efforts are crucial, as awareness and uptake of the vaccine remain low among adolescents and young adults, particularly in the United States. Education and targeted efforts are needed to improve awareness and vaccination rates among this population. The vaccine is particularly important for preventing invasive meningococcal disease, which can have severe consequences. Overall, effective vaccination strategies are essential to reduce the incidence of meningococcal disease.\\
\bottomrule
\end{tabular}
\vspace{-1mm}
\caption{Examples of summaries generated by all the baselines.}
\label{tab:example_baseline}
\end{table*}
\end{document}